\documentclass[runningheads]{llncs}
\usepackage{graphicx}

\usepackage{tikz}
\usepackage{comment}
\usepackage{amsmath,amssymb} %
\usepackage{color}

\usepackage{parallel}
\usepackage{longtable}
\usepackage{dirtytalk}

\usepackage[accsupp]{axessibility}  %

\usepackage{epsfig}
\usepackage{graphicx}
\usepackage{wrapfig}
\usepackage{subcaption}

\usepackage{hyperref}
\usepackage{cleveref}
\usepackage{caption}
\usepackage{textcomp}
\usepackage{stmaryrd}
\usepackage{upgreek}
\usepackage{bm}
\usepackage{cases}
\usepackage{mathtools}
\usepackage{multirow}
\usepackage{bbm}
\usepackage{romannum}

\usepackage{paralist}
\usepackage{enumitem}

\usepackage{cite}

\usepackage{multirow}
\usepackage{rotating}
\usepackage{booktabs}
\usepackage{tabularx}
\usepackage{tabulary}
\usepackage{xcolor,colortbl}

\usepackage{pifont}

\usepackage[linesnumbered,ruled,vlined]{utils/algorithm2e}
\belowdisplayskip \abovedisplayskip

\newcommand{\csection}[1]{
    \vspace{-0.13in}
    \section{#1}
    \vspace{-0.13in}
}

\newcommand{\csubsection}[1]{
    \vspace{-0.12in}
    \subsection{#1}
    \vspace{-0.10in}
}

\newcommand{\cmark}{\ding{51}}%
\newcommand{\xmark}{\ding{55}}%

\newenvironment{compitemize} {
\begin{itemize}[noitemsep,topsep=0pt,parsep=0pt,partopsep=0pt,leftmargin=*] 
}   {\end{itemize}}

\newenvironment{compenumerate} {\begin{enumerate}[noitemsep,topsep=0pt,parsep=0pt,partopsep=0pt,leftmargin=*]}   {\end{enumerate}}

\newcommand{\confint}[1]{\scriptsize $\pm$ #1 }

\usepackage{inconsolata}

\newcommand{\tname}{Housekeep\xspace}

\newcommand{\recmodels}{$395$\xspace}
\newcommand{\reccats}{$32$\xspace}

\usepackage{amsmath,amsfonts,bm}

\def\eqref#1{equation~\ref{#1}}

\def\1{\bm{1}}

\newcommand{\test}{\mathcal{D_{\mathrm{test}}}}

\DeclareMathAlphabet{\mathsfit}{\encodingdefault}{\sfdefault}{m}{sl}
\SetMathAlphabet{\mathsfit}{bold}{\encodingdefault}{\sfdefault}{bx}{n}

\def\gO{{\mathcal{O}}}

\def\gR{{\mathcal{R}}}
\def\gS{{\mathcal{S}}}

\makeatletter
\DeclareRobustCommand\onedot{\futurelet\@let@token\@onedot}
\def\@onedot{\ifx\@let@token.\else.\null\fi\xspace}

\def\eg{\emph{e.g}\onedot} 
\def\ie{\emph{i.e}\onedot}

\makeatother

\makeatletter
\newcommand{\bfgreek}[1]{\bm{\@nameuse{#1}}}
\makeatother

     {\begin{list}{}%
             {\setlength{\leftmargin}{#1}}%
             \item[]%
     }
{\end{list}}

\newcommand{\ttE}{\texttt{E}}

\renewcommand{\KwSty}[1]{\textnormal{\textcolor{blue!90!black}{\ttfamily #1}}\unskip}
\renewcommand{\ArgSty}[1]{\textnormal{\ttfamily #1}\unskip}
\SetKwComment{Comment}{\color{green!50!black}// }{}
\renewcommand{\CommentSty}[1]{\textnormal{\ttfamily\color{green!50!black}#1}\unskip}

\newcommand{\var}{\texttt}

\SetKwProg{Function}{def}{}{}
\renewcommand{\ProgSty}[1]{\texttt{ #1}}
\SetKwProg{Class}{class}{}{}

\SetKwProg{While}{while}{}{}
\SetKwProg{For}{for}{}{}
\SetKwProg{If}{if}{}{}
\SetKwProg{Else}{else}{\unskip}{}

\newcommand{\STAB}[1]{\begin{tabular}{@{}c@{}}#1\end{tabular}}

\usepackage{booktabs}       %
\usepackage{amsfonts}       %
\usepackage{nicefrac}       %
\usepackage{microtype}      %
\usepackage{colortbl}
\usepackage{placeins}

\usepackage{float}
\usepackage{fontawesome}
\usepackage{adjustbox}
\newcolumntype{R}[2]{%
    >{\adjustbox{angle=#1,lap=\width-(#2)}\bgroup}%
    l%
    <{\egroup}%
}

\usepackage{rotating}

\definecolor{Gray}{gray}{0.95}
\newcolumntype{a}{>{\columncolor{Gray}}c}

\begin{document}
\pagestyle{headings}
\mainmatter
\def\ECCVSubNumber{1790}  %

\title{\tname: Tidying Virtual Households using Commonsense Reasoning} %

\author{
    Yash Kant\inst{1,2}\thanks{Work done partially when visiting Georgia Tech} \and
    Arun Ramachandran\inst{2} \and
    Sriram Yenamandra\inst{2} \and
    Igor Gilitschenski\inst{1} \and
    Dhruv Batra\inst{2,3} \and
    Andrew Szot \inst{2}$^\dag$ \and
    Harsh Agrawal \inst{2}$^\dag$
}

\authorrunning{Kant et al.}
\institute{
$^{1}$University of Toronto, $^{2}$Georgia Tech, $^{3}$Meta AI \\
$^\dag$ Equal Contribution
}

\maketitle
\vspace{-15pt}
\begin{abstract}

We introduce \textbf{\tname}, a benchmark to evaluate commonsense reasoning in the home for embodied AI.
In \tname, an embodied agent must tidy a house by rearranging misplaced objects \emph{without explicit instructions specifying which objects need to be rearranged.}
Instead, the agent must learn from and is evaluated against human preferences of which objects \emph{belong} where in a tidy house.
Specifically, we collect a dataset of where humans typically place objects in tidy and untidy houses constituting 1799 objects, 268 object categories, 585 placements, and 105 rooms. 
Next, we propose a modular baseline approach for \tname that integrates planning, exploration, and navigation. 
It leverages a fine-tuned large language model (LLM) trained on an internet text corpus for effective planning. 
We show that our baseline agent generalizes to rearranging unseen objects in unknown environments. See our webpage for more details: \url{https://yashkant.github.io/housekeep/}
\end{abstract}
\vspace{-15pt}

\csection{Introduction}
\begin{figure*}[t!]
    \centering
    \includegraphics[width=0.85\textwidth]{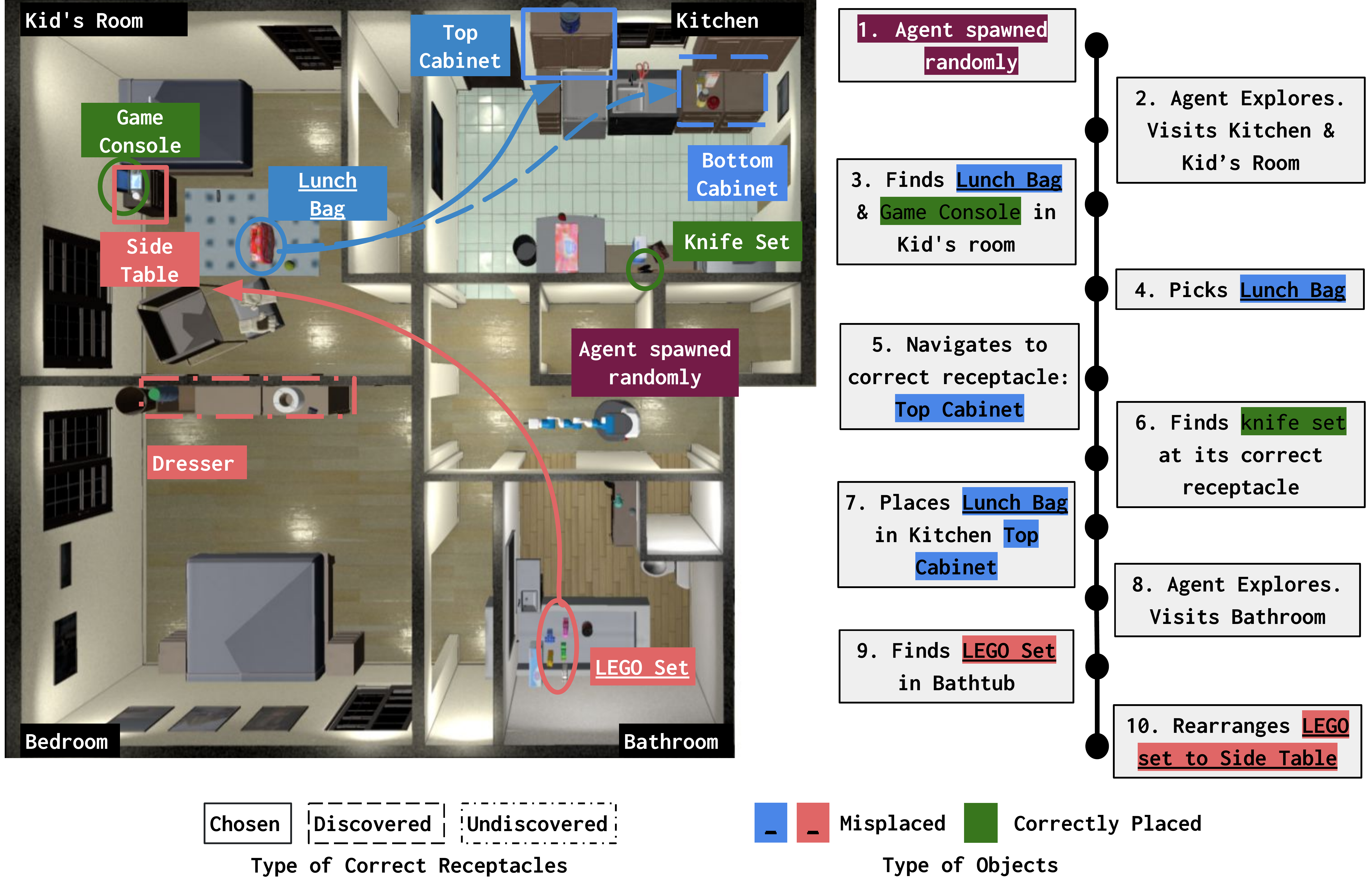}
    \caption{
    \small
      In \tname, an agent is spawned in an untidy environment and tasked with rearranging objects to suitable locations without explicit instructions. 
      The agent explores the scene and discovers misplaced objects, correctly placed objects, and receptacles where objects belong.
      The agent rearranges a misplaced object (like a lunch box on the floor in the kid's room) to a better receptacle like the top cabinet in the kitchen. 
  }
    \label{fig:teaser}
    \vspace{-10pt}
\end{figure*}

Imagine your house after a big party: there are dirty dishes on the dining table, cups left on the couch, and maybe a board game lying on the coffee table. 
Wouldn't it be nice for a household robot to clean up the house \emph{without needing explicit instructions specifying which objects are to be rearranged}? 

Building AI reasoning systems that can perform such housekeeping tasks is an important scientific goal that has seen a lot of recent interest from the embodied AI community. 
The community has recently tackled various problems such as navigation~\cite{anderson2018evaluation, roomnav, wani2020multion, batra2020objectnav, ai2thor, Gan2020ThreeDWorld}, interaction and manipulation~\cite{szot2021habitat, mthor}, instruction following~\cite{anderson2018vision, shridhar2020alfred}, and embodied question answering~\cite{gordon2018iqa,Das2018EmbodiedQA,wijmans2019embodied}.
Each of these tasks defines a goal, e.g. navigating to a given location, moving objects to correct locations, or answering a question correctly.
However, defining a goal for tidying a messy house is more tedious -- one will have to write down a rule for where every object can or cannot be kept. Previous works in semantic reasoning frameworks for physical and relational commonsense~\cite{daruna2019robocse, bisk2019piqa, Bosselut2019, liu2021structformer, LiuRSS2021, Abdo2015RobotOM} are often limited to specific settings (\eg evaluating multi-relational embeddings) without instantiating these tasks in a physically plausible scenario, or by not capturing the full context of a complete household (\eg table-top organization). 
We believe the time may be right to bridge the gap between the above two lines of research. 

We introduce the \tname task to benchmark the ability of embodied AI agents to use physical commonsense reasoning and infer rearrangement goals that mimic human-preferred placements of objects in indoor environments.
\Cref{fig:teaser} illustrates our task, where the Fetch robot is randomly spawned in an unknown house that contains unseen objects.
Without explicit instructions, the agent must then discover objects placed in the house, classify the misplaced ones (LEGO set and lunch bag in \Cref{fig:teaser}), and finally rearrange them to one of many suitable receptacles (matching color-coded square).
We collect a dataset of human preferences of object placements in tidy and untidy homes and use this dataset for: a) generating semantically meaningful initializations of unclean houses, and b) defining evaluation criteria for what constitutes a clean house.
This dataset contains rearrangement preferences for 1799 objects, in 585 placements, in 105 rooms, constituting 1500+ hours of effort from 372 total annotators with 268 object categories curated from the Amazon-Berkeley~\cite{collins2021abo}, YCB objects~\cite{alli2015TheYO}, Google Scanned Objects~\cite{gso}, and iGibson~\cite{shen2020igibson} datasets. 
\tname evaluates how an agent is able to rearrange novel objects not seen during training.

\tname is a challenging task for several reasons. 
First, agents need to reason about the correct placement of \emph{novel} objects.
Second, agents in \tname must operate in unseen environments using only egocentric visual observations. 
We report systematic generalization on unseen houses because we evaluate learning-based techniques.
In the absence of any goal specification, the agent must \emph{explore} areas that get cluttered frequently (\eg coffee table, kitchen counter) for discovering potentially misplaced objects, and also find their suitable receptacles. 
Finally, since the environment is partially observable, the agent must continuously re-plan for when and where to rearrange objects via commonsense reasoning. 
For instance, on discovering a toy on the coffee table in the living room, the agent may choose to not rearrange it immediately if it hasn't discovered a more suitable receptacle such as the closet in the kid's room yet. 
The agent also has to reason about multiple potentially correct receptacles for any given object. For example, a toy could go in the closet in the master bedroom or in the kid's room. 

We propose a modular baseline and demonstrate that embodied (physical) commonsense extracted from large language models (LLMs)~\cite{Brown2020LanguageMA, Liu2019} serves as an effective planner for \tname.
Specifically, we find that finetuning LLM embeddings on a subset of human preferences generalizes well, and helps to reason about correct rearrangements for novel objects never seen during training.
We integrate this LLM-based planning module into a hierarchical policy that coordinates navigation, exploration, and planning as a baseline approach to \tname. Our hierarchical approach also generalizes to unseen objects and scenes in \tname achieving an object success rate of 0.23 for unseen (versus 0.30 on seen objects).
We also qualitatively analyze different failure cases of our baseline to highlight venues for further progress.

\csection{Related Work}

\begin{table*}[t]
	\caption{
    \small
    Comparison of \tname to other rearrangement benchmarks}
    \vspace{-10pt}
    \texttt{
    \begin{center}
        \scriptsize
        \setlength{\tabcolsep}{3pt}
		\begin{tabular}{c l c c c c c c}
			\toprule
			\# & Benchmark	& Goal 	& \begin{tabular}{@{}c@{}}Object\\categories\end{tabular}	& \begin{tabular}{@{}c@{}} Object\\models\end{tabular} & Scenes & Rooms	& Annotators \\
			\midrule
			1 & Transport Challenge~\cite{Gan2020ThreeDWorld}	& Geometric & 50 & 112 & \textbf{15} & 90-120 & - \\
			2 & Habitat 2.0~\cite{szot2021habitat}	& Geometric	& 41 & 92 & 1 & 111 & - \\
            3 & Behavior~\cite{Srivastava2021BEHAVIORBF} &	Predicate &	\textbf{391} & 1217 & \textbf{15} & 100 & - \\
            4 & VRR~\cite{weihs2021visual} & Episodic & 118 & 118 & - & \textbf{120} & -\\
            5 & Taniguchi et al.~\cite{Taniguchi2021AutonomousPB} & Episodic & 55 & 55 & 1 & 4 & - \\
            6 & Jiang et al.~\cite{Jiang2012LearningOA} & Human Preferences & 19 & 47 & - & 20 & 3-5 \\
            7 & My House, My Rules~\cite{Kapelyukh2021MyHM} & Human Preferences & 12 & 12 & 2 & - & 75 \\
            8 & \textbf{\tname}  & Human Preferences & 268 & \textbf{1799} & 14 & 105 & \textbf{372} \\
            \bottomrule
		\end{tabular}
	\end{center}
	}
	\label{tab:task:benchmarks}
	\vspace{-1em}
\end{table*}

\noindent \textbf{Embodied AI Tasks}.
In recent times, we have seen a proliferation of Embodied AI tasks. Benchmarks on indoor navigation use point-goal specification~\cite{habitat19iccv, habitat-challenge}, object-goal~\cite{batra2020objectnav, wani2020multion}, room navigation~\cite{roomnav}, and language-guided navigation~\cite{anderson2018vision, Thomason2019VisionandDialogN}. Some interactive tasks study the agent's ability to follow natural language instruction such as ALFRED~\cite{shridhar2020alfred} and TEACh~\cite{Padmakumar2021TEAChTE} while others focus on rearranging objects following a geometric goal or predicate based specification~\cite{weihs2021visual, Gan2020ThreeDWorld, szot2021habitat, Srivastava2021BEHAVIORBF}. \cite{batra2020rearrangement} provides a summary of rearrangement tasks. All these tasks require an explicit goal specification lifting the burden of learning semantic compatibility of objects and their locations in the house from the agent. In contrast, in this work, we argue that agents shouldn't require an explicit goal specification to perform household tasks such as tidying up the house. Instead, it should use its common sense knowledge to infer the human-preferred goal state.

\textbf{Capturing Human Preferences.} 
Several works (summarized in \cref{tab:task:benchmarks}) in robotics model human preferences for assistive robots.
Some \cite{Jiang2012LearningOA} looked at furniture rearrangement based on surrounding human activities (\eg standing by the kitchen shelf) while others\cite{Kapelyukh2021MyHM, Abdo2015RobotOM} looked at table-top or a shelf rearrangement conditioned on a user. We differ from these works because we are interested in tidying up \emph{entire houses} instead of a particular shelf or a table-top. In addition, the agent needs to operate with partial observations, and generalize to unseen environments and object types. \cite{Taniguchi2021AutonomousPB} comes closest to our work. They learn a spatial model of object placements in a tidy environment. Our benchmark has a larger scale ($1799$ objects spanning $268$ categories vs $\le55$ object instances; $100+$ room configurations vs $1$ scene in ~\cite{Taniguchi2021AutonomousPB}). Our benchmark also tests generalization to \emph{unseen} objects, utilizing a dataset of human preferences instead of learning from a small set of tidy house instances. Dealing with unseen objects is important for real applications since humans can bring new objects into the home. 

\textbf{Commonsense Reasoning}.
Prior work in Natural Language Processing has studied the problem of imbuing commonsense knowledge in AI systems, from social common-sense knowledge~\cite{Levesque2011TheWS,Sap2019, Bosselut2019, Sap2019ATOMICAA, Zellers2019FromRT, Sakaguchi2019WINOGRANDEAA} to understand the likely intents, goals, and social dynamics of people, abductive commonsense reasoning~\cite{Bhagavatula2020AbductiveCR}, next event prediction~\cite{Zellers2019HellaSwagCA, Zellers2018SWAGAL}, to temporal common sense knowledge about temporal order, duration, and frequency of events~\cite{Zhou2019GoingOA,Agrawal2016SortSS,Mostafazadeh2016ACA,GranrothWilding2016WhatHN}. Most similar to our work is the study of physical commonsense  knowledge~\cite{bisk2019piqa} about object affordances, interaction, and properties (such as flexibility, curvature, porousness). However, these benchmarks are static in nature (as a dataset of textual or visual prompts). Our task, on the other hand, is instantiated in an embodied interactive environment and more realistic -- the environment is partially observed, and the agent has to explore unseen regions, discover misplaced objects and use common-sense reasoning to infer compatibility between objects and receptacles. 

\textbf{Application of Large Language Models}.
With the introduction of Transformer~\cite{vaswani2017attention} style architectures, we have seen a diverse range of applications of large language models (LLMs) pre-trained on web-scale textual data. They have not only performed well on natural language processing tasks~\cite{Liu2019, vaswani2017attention}, but the implicit knowledge learned by these models have shown to be effective for other unrelated tasks~\cite{Lu2021PretrainedTA}. LLMs has had a lot of success in vision-and-language tasks like Visual Question Answering (VQA)~\cite{Lu2019ViLBERTPT, Wang2021VLMoUV} and image captioning~\cite{Hu2020VIVOSH, Li2020OscarOA}, external knowledge-based question answering~\cite{Roberts2020HowMK,Brown2020LanguageMA} and construction~\cite{Bosselut2019}. They have also been shown to improve performance on Embodied AI tasks like vision-and-language navigation~\cite{Majumdar2020ImprovingVN, moudgil2021soat}, instruction following~\cite{Hill2020HumanIW}, and planning for embodied tasks~\cite{Li2022PreTrainedLM,Huang2022LanguageMA}. In our work, we explore if language models can display common-sense knowledge of how humans prefer to tidy up their homes.

\csection{\tname: Task and Dataset}
In this section, we will formally define the \tname task and its instantiation in the Habitat~\cite{habitat19iccv, szot2021habitat} simulator. \smallskip

\csubsection{Task Specification}
\label{sssec:task}
\noindent \textbf{Definition}: Recall, in \tname an embodied agent is required to clean up the house by rearranging misplaced objects to their correct location within a limited number of time steps. The agent is spawned randomly in an unseen environment and has to explore the environment to find misplaced objects and put them in their correct locations (receptacles). 

\textbf{Scenes and Rooms}: We use 14 interactive and realistic iGibson scenes~\cite{shen2020igibson}.  These scenes span 17 room types (\eg living room, garage) and contain multiple rooms with an average of 7.5 rooms per scene. We remove one scene from the original iGibson dataset (\var{benevolence\_0\_int}) because it's unfurnished.

\textbf{Receptacles}: We define \emph{receptacles} as flat horizontal surfaces in a household (furniture, appliances) where objects can be found -- misplaced or correctly placed. 
 We remove assets that are neither objects nor receptacles (\eg windows, paintings, etc) and end up with \recmodels unique receptacles spread over \reccats categories. An iGibson scene can contain between 19-78 receptacles.
Notice that a valid object-receptacle placement requires the additional context of what room the receptacle is situated in. 
For example, a counter in the kitchen is a suitable receptacle to place a fruit basket, however, a counter in the bathroom may not be. Hence, we care about the diversity in combinations of room-receptacle occurrences for \tname. Overall, there are 128 distinct room-receptacles in the iGibson scenes.

\textbf{Objects}: We collect objects from four popular asset repositories -- Amazon Berkeley Objects \cite{collins2021abo}, Google Scanned Objects~\cite{gso}, ReplicaCAD~\cite{szot2021habitat}, and YCB Objects~\cite{calli2015ycb}. We filter out objects with large dimensions (\eg ladders, televisions), and objects that do not usually move in a household (\eg garbage cans). After filtering, we have 1799 unique objects spread across 268 categories. We further categorize these objects into 19 high-level semantic categories such as stationery, food, electronics, toys, etc. More details about the filtering, semantic classes, and high/low-level object categories are in the Appendix~\ref{sec:supp:data_stat}.

\textbf{Agent}: 
We simulate a Fetch robot~\cite{fetchrobot}, which has a wheeled base with a 7-DoF arm manipulator, parallel-jaw gripper, and an RGBD camera ($90^{\circ}$ FoV, $128 \times 128$ pixels) on the robot's head. 
The robot moves its base and head through five discrete actions -- move forward by 0.25m, rotate base right or left by 10$^{\circ}$, rotate head camera up or down (pitch) by 10$^{\circ}$.
The robot interacts with objects through a ``magic pointer abstraction'' \cite{batra2020rearrangement} where at any step the robot can select a discrete ``interact'' action.
When invoked, this action casts a ray 1.5m in front of the agent.
If the agent is not currently holding an object and this ray intersects with a graspable object, then the object is now ``held'' by the agent.
If the agent is already holding an object and the ray intersects with a receptacle, then the object is placed on that receptacle. 
Rather than place the object at the point selected on the receptacle, the object is automatically placed on the receptacle.

\csubsection{Human Preferences Dataset: Where Do Objects Belong?}

The central challenge of \tname is understanding how humans prefer to put everyday household objects in an organized and disorganized house. We want to capture where objects are typically found in an unorganized house (before tidying the house), and in a tidy house where objects are kept in their correct position (after the person has tidied the house).  To this end, we run a study on Amazon MTurk~\cite{crowston2012amazon, salganik_bit_2017} with 372 participants. Each participant is shown an object (\eg salt-shaker), a room (\eg kitchen) for context, and asked to classify all the receptacles present in the room into  the following categories: 
\begin{compitemize}
  \item \var{misplaced}: subset of receptacles where object is found \emph{before} housekeeping.
  \item \var{correct}: subset of receptacles where object is found \emph{after} housekeeping. 
  \item \var{implausible}: subset of receptacles where object is unlikely to be found either in a clean or an untidy house.
\end{compitemize}
We also ask each participant to rank receptacles classified under \var{misplaced} and \var{correct}.
For example, given a can of food, someone may prefer placing it in kitchen cabinets while others will rank pantry over the kitchen cabinet.

For each object-room pair ($268\times17$), we collect 10 human annotations.  We collect human annotations through multiple batches of smaller annotation tasks. In a single annotation task, we ask participants to classify-then-rank receptacles for 10 randomly sampled object-room pairs. 
On average a participant took 21 minutes  to complete one annotation task.
Overall, participants spent 1633 hours doing our study. 
\Cref{sec:supp:amt_study} provides more details about the instructions page, user interface, training videos, and FAQs provided in the beginning of the task. \smallskip

\begin{figure}[t]
  \centering
   \begin{subfigure}[t]{0.40\columnwidth}
    \includegraphics[width=\textwidth]{figures/fleiss_objects_hist.pdf}
    \caption{
      \small
      Object Category Agreement
    }
    \label{fig:agreement_per_object}
  \end{subfigure}
  \begin{subfigure}[t]{0.40\columnwidth}
    \includegraphics[width=\textwidth]{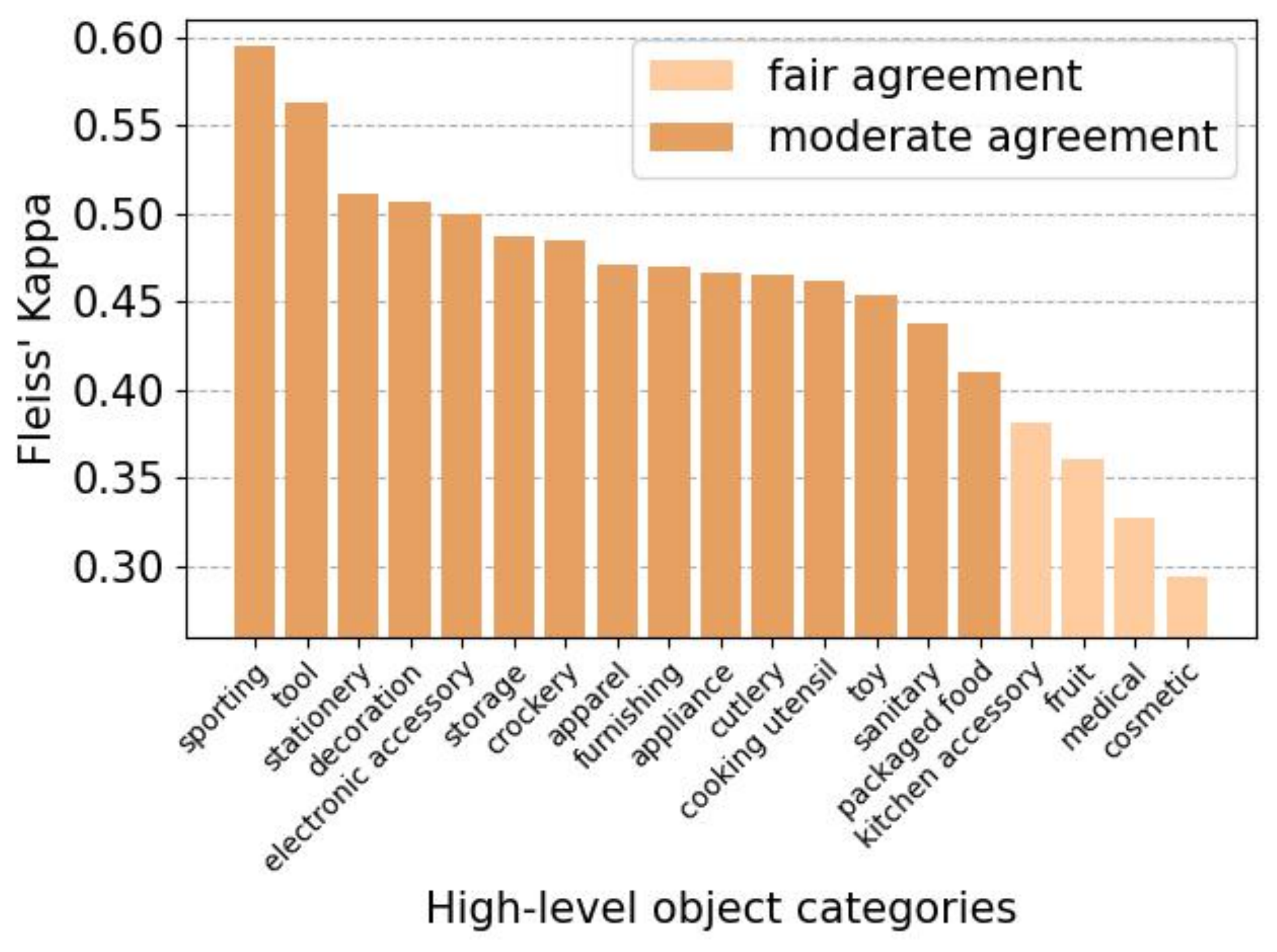}
    \caption{
      \small
      High-Level Category Agreement
    }
    \label{fig:agreement_per_high_level}
  \end{subfigure}
  \caption{
    \small
    Analysis of agreement between reviewer ratings in the \tname human rearrangement preferences dataset.
  }
\end{figure}

\noindent\textbf{Agreement analysis.} 
We evaluate the quality of our human annotations, 
using the Fleiss' kappa (\var{FK}) metric \cite{fleiss1971mns}, which is widely used to assess the reliability of agreement between raters when classifying items. Recall that we collect 10 annotations to classify receptacles for each object-room pair into \var{correct}, \var{misplaced}, or \var{incompatible} bins. In Figure~\ref{fig:agreement_per_object}, we report \var{FK} agreement per object across all room-receptacle pairs ($269 \times 128$) after keeping $8/10$ annotations with the highest inter-human agreement.
We use the agreement ranges proposed by \cite{landis1977measurement} to interpret the \var{FK} scores. We also show agreement when combining the \var{misplaced} and \var{implausible} categories. 
\Cref{fig:agreement_per_object} demonstrates about $90\%$ of our collected data has fair to moderate agreement between annotators.
\Cref{fig:agreement_per_high_level} shows the mean agreement for high-level semantic categories. The agreement is higher for sporting, tool, and stationery categories  because they go to specific places (office desks, garage, etc). The agreement is low for objects like fruits, medicines, packaged foods because people differ in where they like to keep these objects (packaged food can go in cabinets, shelves, kitchen counters, refrigerators).
Overall, these results indicate that our data defines a high-quality source of ground truth rearrangement preferences agreed upon by the majority of annotators.

\csubsection{Episodes}
\label{sec:task:episodes} 
Each \tname episode is created by instantiating 7-10 objects within a scene, out of which 3-5 objects are misplaced and the remaining are placed correctly. 
Next, we concretely define the notions of \emph{correct} and \emph{misplaced} objects. For a given scene, let 
$\gR$
be the set of receptacles available, and
$\gO$
be the set of all the objects which could be instantiated on these. Given an object $o \in \gO$, let $c_{or}$, $m_{or}$ respectively be the ratio of annotators who placed receptacle $r \in \gR$ in \var{correct} and \var{misplaced} bins respectively. We call an object \emph{correctly placed} if $c_{or} > 0.5$, and \emph{misplaced} if $m_{or} > 0.5$, where both cannot be simultaneously true.  \smallskip

\noindent \textbf{Splits}: We create three non-overlapping sets of objects -- \var{seen} (fork, gloves, etc.), \var{val-unseen} (chopping board, dishtowel, etc.), and \var{test-unseen} (banana, scissors, etc.). \var{seen}, \var{val-unseen} and \var{test-unseen} contains 8, 2 and 9 high-level object categories respectively. Note that \emph{only} 40\% of all objects are provided for training, making \tname a strong benchmark to test generalization to unseen objects.      

We also split the 14 scenes into \var{train}, \var{val} and \var{test} with 8:2:4 scenes each respectively. We provide five different splits to test agents on a wide array of commonsense reasoning and rearrangement capabilities.   

\begin{compitemize}
    \item \textbf{train}: 9K episodes with \var{seen} objects and \var{train} scenes
    \item \textbf{val-seen}: 200 episodes with \var{seen} objects and \var{val} scenes
    \item \textbf{val-unseen}: 200 episodes with \var{unseen} objects and \var{val} scenes
    \item \textbf{test-seen}: 800 episodes with \var{seen} objects and \var{test} scenes
    \item \textbf{test-unseen}: 800 episodes with \var{unseen} objects and \var{train} scenes
\end{compitemize} 
\smallskip
More details on episode statistics, and generation are in Appendix~\ref{sec:supp:hk}.

\csubsection{Evaluation}
\label{sec:task:eval} 
We evaluate agents in three different dimensions of rearrangement quality, efficiency, and exploration. All metrics are reported per episode and then aggregated across multiple episodes to report averages and standard errors. While we only describe these metrics informally here, a more nuanced discussion with formal definitions for these can be found in Appendix~\ref{sec:supp:eval} 

\noindent\textit{\textbf{Metrics for Rearrangement.}} These metrics evaluate the relative change in the placement of objects between start and end states of the episode. 
\begin{compitemize}
    \item \textbf{Episode Success (\var{ES})}: Strict binary (\emph{all} or \emph{none}) metric that is one if and only if all objects (irrespective of whether initially misplaced or correctly placed) in the episode are correctly placed at the end of the episode.
    \item \textbf{Object Success (\var{OS})}: Fraction of the objects 
    placed correctly.
  \item \textbf{Soft Object Success (\var{SOS})}: The ratio of reviewers that agree that an object 
  is placed correctly. 
  \item \textbf{Rearrange Quality (\var{RQ})}: A normalized value in $[0, 1]$ (via mean reciprocal rank~\cite{mrr}) is given to each object-receptacle based on the ranking collected from human preferences, $0$ is given if misplaced. 
\end{compitemize}
Metrics \var{OS}, \var{SOS} and \var{RQ} are averaged across objects that are \emph{initially misplaced} or \emph{ever picked up} by the agent during the episode.  

\noindent\textit{\textbf{Exploration and Efficiency Metrics:}} We also study how well the agent explores an unseen environment as well as efficiency at rearranging objects.
\begin{compitemize}
  \item \textbf{Map Coverage (\var{MC})}: The \% of the navigable map area explored. 
  \item \textbf{Misplaced Objects Coverage (\var{MOC})}: The fraction of misplaced objects discovered. Agent discovers an object when it appears in FoV at any point. 
  \item \textbf{Pick and Place Efficiency (\var{PPE})}: The minimum number of picks and places required to solve the episode divided by the number of picks and places made by agent in the episode.
\end{compitemize}

\csection{Methods}
\label{sec:methods} 
In this section, we describe our hierarchical baseline for the \tname benchmark. Our baseline breaks the multi-stage rearrangement into three natural components: a) exploration and mapping, b) planning, and c) navigation and rearrangement. The planning module communicates with all the other modules and determines what the agent does (explore or rearrange). 
Before we dive into the details of our baseline, we discuss some additional sensors that our baseline has access to. 
\noindent\textbf{Additional Sensors}: 
In the \tname specification the agent operates from an RGBD sensor. However, to scope the problem and focus on the planning and commonsense reasoning we allow access to the following: 
\begin{compitemize} 
  \item \emph{semantic} and \emph{instance} sensor: Provides two pixel-wise masks aligned with egocentric RGB observations. The semantic segmentation mask maps every pixel to an object or receptacle category (\eg bowl, cabinet). The instance mask maps every pixel to a unique instance ID, which helps to disambiguate between instances of the same object/receptacle category.
  \item \emph{relationship} sensor: Given instance IDs of an object and a receptacle in the egocentric view, the relationship sensor predicts a binary value if the object is on top of the receptacle or not.
  \item \emph{receptacle-room} map: Receptacles are static within a scene, so we also assume access to a mapping that provides us with the room name for any receptacle discovered (\eg an oven maps to the kitchen). 
\end{compitemize} 
In the future, these sensors can be easily swapped with their learned counterparts.
\cite{jiang2018rednet,cartillier2020semantic} demonstrate it is possible to learn a segmentation sensor for indoor scenes, and \cite{armeni20193d} shows it is possible to learn to infer relationships between 3D objects. 

\csubsection{Mapping and Exploration}

\texttt{\textbf{Mapping:}} At the start of an episode, this module initializes an empty top-down allocentric map. 
As the agent navigates through the environment, it continuously updates the map at each step using egocentric observations and camera projection matrix. We further use the RGBD-aligned pixel-wise instance and semantic masks to localize objects and receptacles and update our allocentric map with them. 
Finally, the mapping module also keeps track of the room and relationship information of discovered objects and receptacles via the \emph{relationship sensor} and known \emph{receptacle-room map}.

\noindent \texttt{\textbf{Exploration:}} 
To discover misplaced objects as well as suitable receptacles to place them on, our exploration module aims to maximize the area on the map it has seen.
This module only requires the hyperparameter \var{n}$_e$ --- the number of exploration steps --- as input and executes low-level actions via the navigation module. 
We use frontier-based exploration~\cite{yamauchi1997frontier} (FRT) for our main experiments, which iteratively visits unexplored frontiers, which are the edges between visited and unvisited space. 
We keep our implementation details same as those in~\cite{ramakrishnan2020exploration}.

\csubsection{Planning}

\begin{wrapfigure}[20]{r}{0.52\textwidth}
\vspace{-50pt}

\scriptsize
\begin{algorithm}[H]
  \caption{\small \label{alg:plan} Planner}

  \textbf{import modules:} rank \texttt{L}; explore \texttt{E}; map \texttt{M}; navigate \texttt{N}; rearrange \texttt{R}; pick-place \texttt{P} \\
  \textbf{variables}: exploration steps \var{n$_e$}; max steps \var{n} 
  
  \SetInd{0.2em}{0.5em}
    
    \Function{plan(t=0):}{
        \While{t < n:~~\var{\textcolor{gray}{\# stop when t=n at any line}}}{
            \var{\textcolor{gray}{\# nothing to rearrange}} \\
            \If{not \texttt{R}.rearrangements():}{
                \var{\textcolor{gray}{\# explore for n$_e$ steps}} \\
                \For{i in range(n$_e$):}{
                    \var{\textcolor{gray}{\# take an exploration step}} \\
                    \var{obs = \ttE.act(M, N)} \\
                    \var{\textcolor{gray}{\# update map and rearrange modules}} \\
                    \var{M.update(obs); R.update(obs)} \\
                }
                \var{t = t + n$_e$} \\
                \var{R.rescore(L)}~~\var{\textcolor{gray}{\# update scores using L}}
            }
            \Else{:}{
                \var{\textcolor{gray}{\# rearrange until finished}} \\
                \For{r in R.rearrangements():}{
                    \var{\textcolor{gray}{\# object and correct receptacle}} \\
                    \var{obj,rec = r.obj,r.rec} \\
                    \var{\textcolor{gray}{\# nav \& pick obj, then nav \& place on rec}} \\
                    \If{N.nav(obj) \& P.pick(obj) \& N.nav(rec) \& P.place(obj, rec):}{\var{M.update(obs); R.update(obs)}}
                    \var{t = t+n$_r$} \var{\textcolor{gray}{\# update steps}} \\
                }
            }
        }
    }
    
\end{algorithm}

\end{wrapfigure}

Our planner communicates with all the modules to build a high-level rearrangement plan that the agent follows. It consists of:

\noindent\textbf{\var{Rearrange} submodule}: Stores a list of locations of discovered objects and receptacles. 
From this list, it produces a list of object-receptacle pairs indicating the order of rearrangements to perform.
There are 3 key decisions the rearrange submodule needs to make to create this list: 1) what objects are misplaced, 2) what order to arrange misplaced objects, and 3) what receptacle to place each misplaced object on.
It makes these decisions via a \var{\textbf{Ranker}} submodule which ranks potential object-receptacle pairings by modeling the joint distribution $ \mathbb{P} ( \text{receptacle}, \text{room} | \text{object})$.
To solve (3), for a given object the agent picks the receptacle in the room with the highest joint probability.
We model the joint distribution of the receptacle and room because the context of a receptacle will change based on the room.
For example, a plate belongs on the counter in the kitchen, but not a counter in the bathroom.
\Cref{sec:approach:ranking} describes how we compute $ \mathbb{P} ( \text{receptacle}, \text{room} | \text{object})$, and also how we solve (1).
To solve (2), we evaluate 4 heuristic orderings which are described in \Cref{ssec:supp:plan_abl}.

\noindent \textbf{\var{Planner} submodule}: At any given step, the planner decides to explore only if there are no more pending rearrangements. The agent explores for a fixed number of steps (\var{n}$_e$). Intuitively, higher values of \var{n}$_e$ will encourage the agent to explore the environment at the beginning of the episode whereas lower values of \var{n}$_e$ will encourage the agent to rearrange as soon as a better receptacle is found. While exploring, the planner ensures that map and rearrange modules are synchronized at each step. At the end of the exploration phase, the planner uses the rank (\var{L}) module to update compatibility scores by considering newly discovered objects and receptacles. We provide the planner pseudocode in Algorithm~\ref{alg:plan}.

\noindent \textbf{Navigation and Pick-Place}: Please see \Cref{sec:supp:navpp} for details.

\csubsection{Extracting Embodied Commonsense from LLMs}
\label{sec:approach:ranking} 

One of the main goals of \tname is to equip the agent with commonsense knowledge to reason about the compatibility of an object with different receptacles present across different rooms. 
Large Language Models (LLMs) trained on unstructured web-corpora have been shown to work well for several embodied AI tasks like navigation~\cite{Majumdar2020ImprovingVN,hong2020recurrent, Hill2020HumanIW, Li2022PreTrainedLM, Huang2022LanguageMA}. We study whether we can use LLMs to extract physical (embodied) common sense about how humans prefer to rearrange objects to tidy a house.  For this, we build a ranking module 
(\texttt{L})
which takes as input a list of objects and a list of receptacles in rooms and then outputs a sequence of desired rearrangements based on which object receptacle pairings are most likely.
We select the rearrangements that maximize $ \mathbb{P} ( \text{receptacle}, \text{room} | \text{object})$.
We decompose computing this probability into a product of two probabilities:
\begin{compitemize}
\itemsep0em
\item Object Room \var{[OR] -- $\mathbb{P} (\text{room} | \text{object} )$} : Generate compatibility scores for rooms for a given object.
\item Object Room Receptacle \var{[ORR] -- $\mathbb{P} (\text{receptacle} | \text{object}, \text{room} )$}: Generate compatibility scores for receptacles within a given room and for a given object.
\end{compitemize}

Both of these are learned from the human rearrangement preferences dataset. From the compatibility scores in the \var{ORR} task, we first determine which objects in our list of objects are misplaced and which are correctly placed. To do this, we compute a hyperparameter $s_L$ --- the score threshold --- from our \var{val} episodes using a grid search. Receptacles whose scores are above $s_L$ for a given object-room pair are marked as correct, while those whose scores are below $s_L$ are marked as incorrect. We then treat this as a classification task and pick $s_L$ that maximizes the F1 score on the \var{val} episodes.

Next, to determine the ranking of receptacles for a given misplaced object, we use the probabilities from both the \var{OR} and \var{ORR} tasks. For a given object, we first rank the rooms in descending order of $\mathbb{P} (\text{room} | \text{object} )$. Then, for each object-room pair in the ranked room list, we rank the \emph{correct} receptacles in the room in descending order of $\mathbb{P} (\text{receptacle} | \text{object}, \text{room} )$. Finally, we place the \emph{incorrect} receptacles at the end of our list.

To learn the probability scores in the \var{OR} and \var{ORR} tasks, we start by extracting word embeddings from a pretrained RoBERTa LLM~\cite{Liu2019} of all objects, receptacles. 
We experiment with various contextual prompts~\cite{petroni2019language, petroni2020how} for extracting embeddings of paired room-receptacle (\eg \say{\var{<receptacle> of <room>}}) and object-room (\eg \say{\var{<object> in <room>}}) combinations. 
Next, we implemented the following 2 methods of using these embeddings to get the final compatibility scores:

\noindent \textbf{Finetuning by Contrastive Matching (\var{CM}).} We train a 3-layered MLP on top of language embeddings and compute pairwise cosine similarity between any two embeddings.  %
Embeddings are trained using objects from \var{seen} split. We train separate models for \var{ORR} and \var{OR}. 
For \var{ORR}, we match an object-room pair to the receptacle with the best average rank across annotators. We use contrastive loss~\cite{oord2018representation} to promote similarity between an object-room pair and the matching receptacle. 
For \var{OR}, we match an object with all rooms that have at least one \var{correct} receptacle for it. In this case, we use the binary cross entropy (BCE) loss to handle multiple rooms per object. 

\noindent \textbf{Zero-Shot Ranking via MLM (\var{ZS-MLM}).} Masked Language Modeling (MLM) is used extensively for pretraining LLMs~\cite{Liu2019, Devlin2018}, which involves predicting a masked word (\ie \var{[mask]}) given the surrounding context words. This objective can be extended for zero-shot ranking using various contextual prompts. For \var{ORR}, we use the prompt \say{\var{in <room>, usually you put <object> <spatial-preposition> [mask]}} to rank receptacles given an object, a room, and a spatial preposition (\eg in or on). For \var{OR}, we use the prompt \say{\var{in a household, it is likely that you can find <object> in the room called [mask]}}. %

We compare these ranking approaches with other baselines in \Cref{sec:exp:llm}. 
We provide training details of our ranking module in \Cref{sec:supp:approach}.

\csection{Experiments}
We first test whether LLMs can capture the embodied commonsense reasoning needed for planning in \tname. Then we deploy our modular agent equipped with this LLM-based planner to benchmark its ability to generalize to unseen environments cluttered with novel objects from seen (\ie \var{test-seen}) and unseen (\ie \var{test-unseen}) categories. Finally, we perform a thorough qualitative analysis of its failure modes and highlight directions for further improvements. 

\subsection{Language Models Capture Embodied Commonsense}
\label{sec:exp:llm} 

\begin{wraptable}[6]{r}{0.5\linewidth}
\vspace{-25pt}
    \caption{\small We report mAP scores on train, and unseen objects splits of val and test for both \var{OR} and \var{ORR} matching tasks. The finetuning with \var{CM} objective is performed using objects \emph{only} from train split}
    \vspace{-10pt}
    \scriptsize
    \begin{center}
		\setlength{\tabcolsep}{2pt}
		\begin{tabular}{ll ccc ccc}
			\toprule
            &  & \multicolumn{3}{c}{\textbf{\var{ORR}}} & \multicolumn{3}{c}{\textbf{\var{OR}}} %
            \\
			\cmidrule(lr){3-5} \cmidrule(lr){6-8} 
            \texttt{\#} & \textbf{\var{Method}} & \var{train} &  \var{val-u} &  \var{test-u} & \var{train} & \var{val-u} & \var{test-u} %
            \\
            \midrule
            \texttt{1} & \texttt{RoBERTa+CM} & 0.81 & \textbf{0.79} & \textbf{0.81} & \textbf{1.0} & \textbf{0.65} & 0.65 
            \\ 
            \texttt{2} & \texttt{GloVe+CM} & \textbf{0.88} & 0.76 & 0.76 & \textbf{1.0} & \textbf{0.65} & \textbf{0.66} 
            \\ 
            \texttt{3} & \texttt{ZS-MLM} &  0.43 & 0.46 & 0.42 & 0.51 & 0.54 & 0.52 
            \\ 
            \texttt{4} &\texttt{Random} 
                       & 0.47%
                       & 0.47%
                       & 0.46%
                       & 0.58%
                       & 0.52%
                       & 0.59%
            \\ 
            \bottomrule
		\end{tabular}
	\end{center}
	\label{tab:orm}
\end{wraptable}

\textbf{Methods.} We evaluate \var{CM} and \var{ZS-MLM} using RoBERTa~\cite{Liu2019} as our base LLM. We also compare these with GloVe-based~\cite{pennington2014glove} embeddings, and a baseline that randomly ranks rooms (for \var{OR} task) and receptacles (for \var{ORR} task).

\noindent \textbf{Evaluation.} We evaluate mean average precision (mAP) across objects to compare the ranked list of rooms/receptacles obtained from our ranking module
to the list of rooms/receptacles deemed \var{correct} by the human annotators. Recall from \cref{sec:task:episodes}, for a given object, a receptacle is considered \var{correct}  when at least 6 annotators vote for it, and a room is considered \var{correct} if it has at least one \var{correct} receptacle within it. Higher AP score indicates \var{correct} items are likely to ranked higher than the \var{incorrect} items.

\noindent \textbf{Results.} \Cref{tab:orm} shows that \var{RoBERTa+CM} outperforms \var{ZS-MLM} by a large margin even when fintuned on a relatively small-sized training set (\url{~}40\% of total data, see Section~\ref{sec:task:eval}). We find good transfer of results from \var{val} to \var{test} splits by \var{RoBERTa+CM} method on both tasks demonstrating the better generalization capabilities of LLMs. Whereas, \var{GloVe+CM} do not seem to transfer well for the \var{ORR} task. Finally, notice that \var{Random} baseline performs relatively well on room-matching (\var{OR}) task, which is expected since there are ample of rooms with at least one correct receptacle for any given object.    
 
\csubsection{Main Results for \tname}
\label{sec:exp:main} 
We utilize the best method from \Cref{sec:exp:llm}, \var{RoBERTa+CM} as scoring function within \var{Ranker} module to continuously rerank (thus replan) newly discovered rooms and receptacles while exploring \tname episodes.

\noindent \textbf{Oracle Modules.} We show oracle agent's performance, by swappping \var{Ranker} and \var{Explore} modules with their oracle (perfect) counterparts.  Oracle ranker uses the ground truth human preferences to rank the objects and receptacles found. Oracle exploration gives a complete map of the environment, \ie agent knows all objects, receptacles and their respective locations.

\begin{table*}[t!]
	\setlength{\tabcolsep}{3pt}
	\caption{
    \small
    Results using our modular baseline on the \tname \var{test-seen} and \var{test-unseen} splits.
    \var{OR}: Oracle, \var{LM}: LLM-based ranking, \var{FT}: Frontier exploration.
        }

	\begin{center}
	\resizebox{1\linewidth}{!}{
		\begin{tabular}{c c l l ca cc cc cc}
			\toprule

			& & \multicolumn{2}{c}{\scriptsize\textbf{Modules}} & \multicolumn{2}{c}{\scriptsize\textbf{Rearrange}} & \multicolumn{2}{c}{\scriptsize\textbf{Soft-Score}}  & \multicolumn{2}{c}{\scriptsize\textbf{Explore}} & \multicolumn{1}{c}{\scriptsize\textbf{Efficiency}}  \\
			\cmidrule(r){3-4} \cmidrule(lr){5-6} \cmidrule(lr){7-8} \cmidrule(lr){9-10} \cmidrule(l){11-11} 
			& \scriptsize \texttt{\#} & \scriptsize Rank & \scriptsize Explore & \scriptsize\textbf{\texttt{ES}}~$\uparrow$ & \scriptsize\textbf{\texttt{OS}}~$\uparrow$ & \scriptsize\textbf{\texttt{SOS}}~$\uparrow$ & \scriptsize\textbf{\texttt{RQ}}~$\uparrow$ &\scriptsize\textbf{\texttt{MC}}~$\uparrow$ &  \scriptsize\textbf{\texttt{OC}}~$\uparrow$ & \scriptsize\textbf{\texttt{PPE}}~$\uparrow$  \\
			\midrule

\multirow{4}{*}{\STAB{\rotatebox[origin=c]{90}{\texttt{\textbf{t-seen}}}}}
& \scriptsize \texttt{1} & \texttt{OR} & \texttt{OR} & 1.00 \confint{0.00} & 1.00 \confint{0.00} & 0.65 \confint{0.00} & 0.63 \confint{0.00} & -- & 1.00 \confint{0.00} & 1.00 \confint{0.00} \\
& \scriptsize \texttt{2} & \texttt{OR} & \texttt{FTR} & 0.35 \confint{0.02} & 0.64 \confint{0.01} & 0.49 \confint{0.01} & 0.41 \confint{0.01} & 73 \confint{1} & 0.73 \confint{0.01} & 1.00 \confint{0.00} \\
& \scriptsize \texttt{3} & \texttt{LM} & \texttt{OR} & 0.04 \confint{0.01} & 0.44 \confint{0.01} & 0.46 \confint{0.00} & 0.30 \confint{0.01} & -- & 1.00 \confint{0.00} & 0.57 \confint{0.01} \\
& \scriptsize \texttt{4} & \texttt{LM} & \texttt{FTR} & 0.01 \confint{0.00} & 0.30 \confint{0.01} & 0.39 \confint{0.00} & 0.19 \confint{0.01} & 77 \confint{1} & 0.76 \confint{0.01} & 0.41 \confint{0.01} \\
\midrule
\multirow{4}{*}{\STAB{\rotatebox[origin=c]{90}{\texttt{\textbf{t-unseen}}}}}
& \scriptsize \texttt{5} & \texttt{OR} & \texttt{OR} & 1.00 \confint{0.00} & 1.00 \confint{0.00} & 0.64 \confint{0.00} & 0.61 \confint{0.00} & -- & 1.00 \confint{0.00} & 1.00 \confint{0.00} \\
& \scriptsize \texttt{6} & \texttt{OR} & \texttt{FTR} & 0.35 \confint{0.02} & 0.65 \confint{0.01} & 0.49 \confint{0.01} & 0.40 \confint{0.01} & 74 \confint{1} & 0.74 \confint{0.01} & 1.00 \confint{0.00} \\
& \scriptsize \texttt{7} & \texttt{LM} & \texttt{OR} & 0.02 \confint{0.00} & 0.32 \confint{0.01} & 0.42 \confint{0.00} & 0.20 \confint{0.01} & -- & 1.00 \confint{0.00} & 0.42 \confint{0.01} \\
& \scriptsize \texttt{8} & \texttt{LM} & \texttt{FTR} & 0.01 \confint{0.00} & 0.23 \confint{0.01} & 0.36 \confint{0.00} & 0.14 \confint{0.01} & 73 \confint{1} & 0.74 \confint{0.01} & 0.35 \confint{0.01} \\

            \bottomrule
		\end{tabular}
		}
	\end{center}
	
	\label{tab:main:unseen}
\vspace{-10pt}
\end{table*}

\noindent \textbf{Upper Bounds.} In \Cref{tab:main:unseen}, we show results on both \var{test-seen} and \var{test-unseen} splits. Rows 1, 5 with oracle ranking and exploration denote the upper bounds achievable across all metrics. Note that Soft Object Success (\var{SOS}) and Rearrangement Quality (\var{RQ}) are not perfect since human agreement across correct receptacles is not 100\%.

\begin{wrapfigure}[12]{r}{0.4\textwidth}
    \vspace{-30pt}
    \centering
    \includegraphics[width=1.1\linewidth]{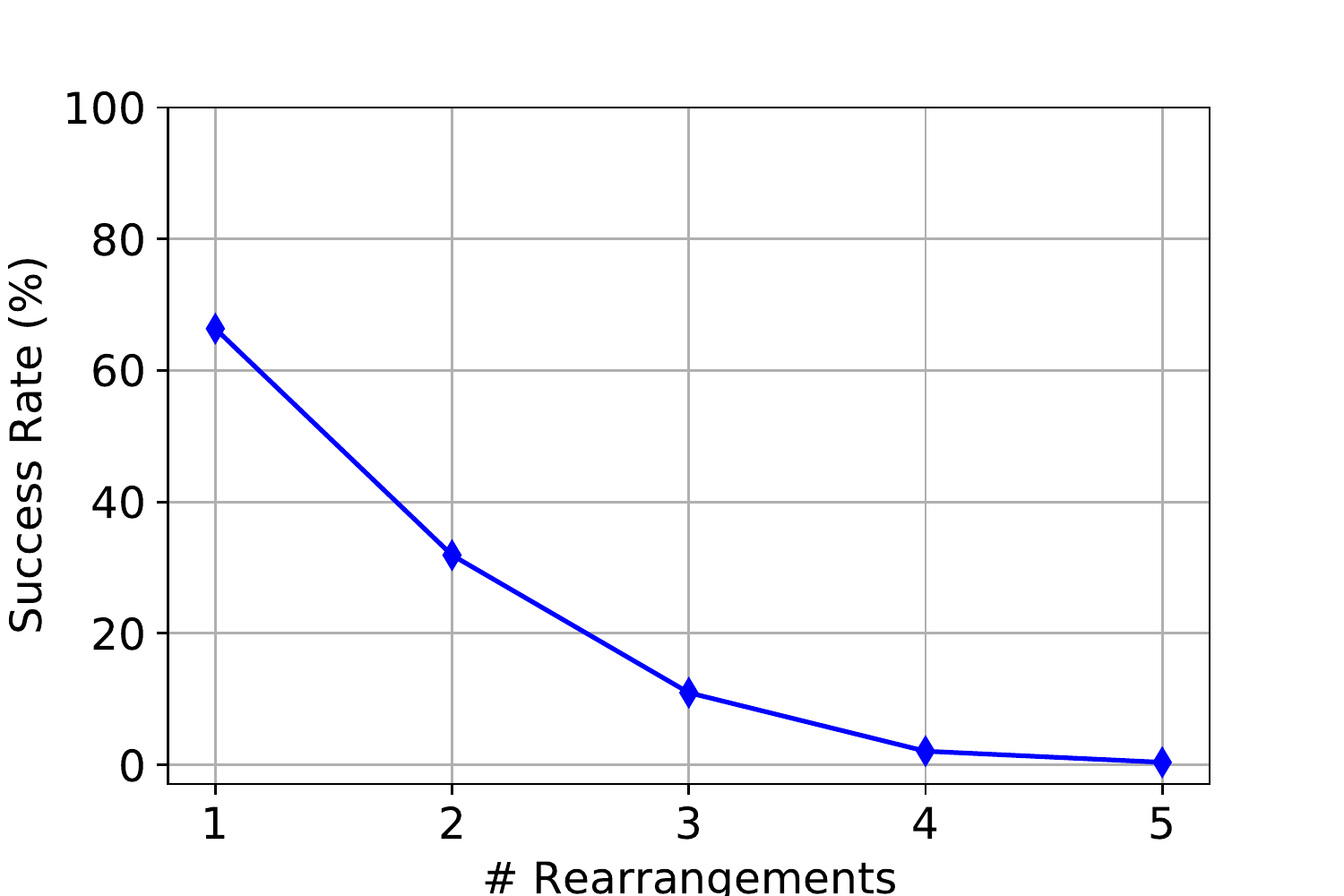}
    \caption{\small Episode Success (\var{ES@K}) vs.\ number of rearrangements (\var{K}) using non-oracle baseline. As \var{K} increases, errors compound, and \var{ES} drops.
  }
    \label{fig:part_succ}
\end{wrapfigure}

\noindent \textbf{LLM-based Ranker, Compounding Errors.} Compared to oracle ranker (Row 1) language model (Row 3) impacts object success (\var{OS}) by -56\%, and episode success (\var{ES}) by -96\%. The dramatic drop in \var{ES} is expected as \tname is a multi-step problem with compounding errors between rearrangements. That means, with average 4 rearrangements necessary per episode and with \var{OS} at $46\%$, \var{ES} will be $0.46^{4} \approx 0.045$ as seen. We further analyze this in \Cref{fig:part_succ} showing that \var{ES@K} drops with each successive rearrangement attempt made.

\noindent \textbf{Frontier Exploration, Full baseline.} Using Frontier exploration (rows 1,2), \var{OS} drops by $47\%$. This drop in performance signifies the importance of task-driven exploration needed for \tname to find \var{misplaced} objects or \var{correct} receptacles quickly. Finally, we evaluate the fully non-oracle baseline (row 4) which achieves a $30\%$ object success rate. From rows 4 and 8, we see that \var{OS} drops by $7\%$, but \var{SOS} drops only by $3\%$ across seen vs unseen objects which supports our claim from Section~\ref{sec:exp:llm} that LLMs can indeed serve as a generalizable planning module aligned with human preferences. 

We put additional experiments analyzing the effect of exploration steps (\var{n}$_e$), exploration strategies in \Cref{sec:supp:add_exps}, and qualitative results in \Cref{sec:supp:qual_analysis}.

\csubsection{Qualitative Analysis}
\label{sec:exp:qual_results}
\begin{figure*}[t!]
    \centering
    \includegraphics[width=1\textwidth]{figures/gantt_75_annotated.pdf}
    \caption{
        \small
        Visually depicting agent's progress on 75 randomly-sampled episodes from two test scenes, \var{beechwood\_1} and \var{benevolence\_1}. Plots (i) and (iii) depict Agent's state, (ii) and (iv) show \% of objects discovered on y-axis, and x-axis is the timestep. State and Discovery plots of same scenes are aligned, \ie show same episodes on Y-axis. 
  }
    \label{fig:gantt}
    \vspace{-5pt}
\end{figure*}

\noindent \Cref{fig:gantt} visually depicts the baseline agent's progress across episodes on two test scenes. Agent State plots show the \emph{module currently being executed}: \var{explore} (blue), \var{rearrange} (orange), or \var{pick/place} (red). Object Discovery plots show the \emph{percentage of misplaced objects discovered} until any given time step. Dark to light shade corresponds to an increasing number of misplaced objects found. Each row corresponds to one episode, and the x-axis denotes time step. 

\noindent \textbf{Agent cannot classify discovered objects as misplaced.}  For \var{beechwood\_1}, row \var{2a} in \textbf{(\romannum{1})} shows that in approximately a quarter of the episodes, the agent only explores and never rearranges.
The corresponding row \var{2b} in \textbf{(\romannum{2})}
tells us that all the misplaced objects were discovered by $\approx$ 500 time steps. From row \var{2a} and \var{2b}, we can conclude that the ranking module fails to identify objects as misplaced even after discovering them.

\noindent \textbf{Agent rearranges incorrect objects.} Next, looking at orange regions in row \var{1a}, we know that the agent rearranges several objects. However, the corresponding row  \var{1b} in \textbf{(\romannum{2})} is fully black, indicating that the agent discovered 0\% of misplaced objects. This means that the reasoning module misidentifies correctly placed objects as misplaced and asks the agent to rearrange them. Moreover, the exploration module fails to locate misplaced objects. 

\noindent \textbf{Scene layouts affect object discovery.} Our agent explores differently in different scene layouts. In \Cref{fig:gantt}, the agent discovers misplaced objects much more quickly in \var{benevolence\_1} than in \var{beechwood\_1}. 
Rows \var{3a} and \var{3b} show this trend -- all objects are discovered within the first 200 steps of the episode in stark contrast to \var{beechwood\_1} episodes. This is explained by the fact that \var{benevolence}\_1 is a smaller home with just one partitioning wall (4 rooms) versus \var{beechwood\_1} (8 rooms)  making exploration and object discovery easier. We also provide top-down maps of both scenes in Appendix~\ref{ssec:supp:topdown}.

\csection{Conclusion}

In this work we presented the \tname benchmark to evaluate commonsense reasoning in the home for embodied AI. We started by collecting a dataset of human preferences of where objects go in tidy and untidy houses, and used it to generate episodes and evaluate agent performance in \tname. Then we proposed a modular and hierarchical baseline that plans using commonsense reasoning extracted from a large language model. We showed this method generalizes to rearranging unseen objects without access to explicit instructions. \tname is a challenging task, and the overall episode success rate remains low despite the use of additional sensors (\eg segmentation, relationship) needed to focus on the planning and commonsense reasoning within the task. Two areas of improvement to our current baseline are our exploration module and reasoning module. With a learned exploration module the agent can visit areas that get cluttered more frequently, and optimize object coverage instead of map coverage. Additionally, improving the reasoning module's recall and precision at identifying misplaced objects can drastically increase performance on our task. Finally, replacing additional sensors related with their learned counterparts will make our baselines more realistic and allow for comparisons with other types of end-to-end learned (\eg RL/IL) policies.

\clearpage
\bibliographystyle{splncs04}
\bibliography{egbib}

\newpage
\appendix

\title{\tname: Appendix}
\author{}
\institute{}
\maketitle

\section{Data Statistics}
\label{sec:supp:data_stat}

In this section we provide details about category level breakdown of objects and receptacles.%

\subsection{High-level Object and Receptacle Categories}
\label{ssec:supp:high_level}
Table~\ref{table:high_level_categories} details the high-level categorization and frequencies of object and receptacles. We also provide one example of every high-level category, and the original source of the data. We gather 2194 object and receptacle models from multiple sources after filtering objects that are not useful for the task. %

\noindent \textbf{Object Filtering Details.} We used category-based filtering for ReplicaCAD, and AB datasets (\eg sofa, bikes, etc) to remove unhelpful objects. Then, we removed objects if any of their dimensions exceeded 50 meters. We also used some manual filtering in order to remove very small objects (\eg keychains).
\begin{table*}[!ht]

    \caption{\small{\textbf{High-level categories.}: This table lists the high-level categories of objects and receptacles and the number of object/receptacle models from each data source for each high-level category}}
    \texttt{
    \resizebox{\textwidth}{!}{
    \begin{tabular}{|c|c|c|cccccc|}
        \hline
        \textbf{High-level category} & \textbf{No. of object}& \textbf{Example} & \multicolumn{6}{c|}{\textbf{No. of models}}\\
        &  \textbf{categories} & & YCB~\cite{alli2015TheYO}&R-CAD~\cite{szot2021habitat}&iGibson~\cite{shen2020igibson}&AB~\cite{collins2021abo}&GSO~\cite{gso}&Total\\
        \hline
        \multicolumn{9}{|c|}{\textit{Objects}}  \\
        \hline
        packaged food & 37 & condiment & 10 & 3 & 0 & 0 &  48 & 61 \\
        fruit & 8 & peach & 8 & 0 & 0 & 0 &  0 & 8 \\
        cooking utensil & 14 & dispensing closure & 3 & 3 & 0 & 4 &  14 & 24 \\
        sanitary & 19 & bath sheet & 2 & 2 & 0 & 1 &  34 & 39 \\
        crockery & 8 & tumbler & 8 & 10 & 0 & 8 &  22 & 48 \\
        cutlery & 6 & plate & 4 & 3 & 0 & 0 &  9 & 16 \\
        tool & 14 & scissors & 11 & 0 & 0 & 0 &  12 & 23 \\
        stationery & 11 & invitation card & 1 & 6 & 0 & 5 &  22 & 34 \\
        sporting & 8 & dumbbell & 6 & 0 & 0 & 27 &  0 & 33 \\
        toy & 36 & video game & 13 & 0 & 0 & 0 &  282 & 295 \\
        electronic accessory & 24 & hard drive & 0 & 1 & 0 & 45 &  95 & 141 \\
        storage & 18 & waste basket & 0 & 2 & 0 & 22 &  33 & 57 \\
        furnishing & 3 & cushion & 0 & 2 & 2 & 222 &  1 & 227 \\
        decoration & 9 & string lights & 0 & 2 & 21 & 59 &  51 & 133 \\
        apparel & 8 & shoe & 0 & 10 & 0 & 2 &  266 & 278 \\
        appliance & 23 & thermal laminator & 0 & 7 & 23 & 215 &  23 & 268 \\
        kitchen accessory & 8 & lime squeezer & 0 & 2 & 0 & 0 &  8 & 10 \\
        medical & 5 & antidepressant & 0 & 0 & 0 & 0 &  66 & 66 \\
        cosmetic & 9 & face moisturizer & 0 & 0 & 0 & 0 &  38 & 38 \\
        \hline
        \multicolumn{9}{|c|}{\textit{Receptacles}}  \\
        \hline
        furniture & 17 & sofa & 0 & 0 & 320 & 0 & 0 & 320 \\
        appliance & 13 & fridge & 0 & 0 & 64 & 0 & 0 & 64 \\
        storage & 2 & basket & 0 & 0 & 11 & 0 & 0 & 11 \\
        \hline
        Total & 268 + 32 & - & 66 & 53 & 441 & 610 & 1024 &  2194 \\
        \hline
    \end{tabular}
    }
    }
    \label{table:high_level_categories}
\end{table*}

\subsection{Low-level Object Categories}
\label{ssec:supp:low_level} 
Table~\ref{table:object_categories} lists the object categories in each of the train, val-unseen and test-unseen splits. The train split has 8 high-level categories, val-unseen has 2 high-level categories and test-unseen split has 9 high-level categories.
\begin{table}[t!]
    \centering
    \scriptsize
    \definecolor{graycell}{rgb}{.85,.85,.85}
    \caption{\small{Object categories in train, val-unseen and test-unseen splits}}
    \texttt{
    \begin{tabular}{  m {0.2cm} | p{3cm} | p{8.5cm}|}
        \multicolumn{1}{c}{} & \multicolumn{1}{c |}{\textbf{High-level category}} & \multicolumn{1}{c}{\textbf{Object categories}} \\
        \hline
        & \textbf{apparel}& cloth, gloves, handbag, hat, heavy duty gloves, helmet, shoe, umbrella\\
        & \cellcolor{graycell}{\textbf{appliance}}& \cellcolor{graycell}{camera, clock, coffeemaker, electric heater, fitness tracker wristband, flashlight, hair dryer, hair straightener, instant camera, lamp, laptop, light bulb, milk frother, portable speaker, router, set-top box, shredder, stand mixer, table lamp, tablet, thermal laminator, toaster, virtual reality viewer}\\
        & \textbf{cooking utensil}& blender jar, bundt pan, casserole dish, dispensing closure, dutch oven, pan, pitcher base, pressure cooker, ramekin, saute pan, skillet, skillet lid, spatula, teapot\\
        \multirow{5}{*}{\rotatebox{90}{\textbf{train}}} & \cellcolor{graycell}{\textbf{cutlery}}& \cellcolor{graycell}{fork, knife, knife block, plate, saucer, spoon}\\
        & \textbf{decoration}& candle holder, lantern, picture frame, plant, plant container, plant saucer, string lights, surface saver ring, vase\\
        & \cellcolor{graycell}{\textbf{medical}}& \cellcolor{graycell}{antidepressant, dietary supplement, laxative, medicine, weight loss guide}\\
        & \textbf{packaged food}& butter dish, cake mix, cake pan, candy, candy bar, cereal, chocolate, chocolate box, chocolate milk pods, chocolate powder, coffee beans, coffee pods, condiment, cracker box, donut, fondant, fruit snack, gelatin box, heavy master chef can, herring fillets, master chef can, mustard bottle, peppermint, pepsi can pack, pet food supplement, potted meat can, pudding box, salt shaker, snack cake, sparkling water, sugar box, sugar sprinkles, tea can pack, tea pods, tomato soup can, water bottle, xylitol sweetener\\
        & \cellcolor{graycell}{\textbf{sporting}}& \cellcolor{graycell}{baseball, dumbbell, dumbbell rack, golf ball, mini soccer ball, racquetball, softball, tennis ball}\\
        \hline
        & \textbf{kitchen accessory}& can opener, chopping board, dish drainer, honey dipper, lime squeezer, spoon rest, sushi mat, utensil holder\\
        \multirow{-2}{*}{\rotatebox[origin=c]{90}{\textbf{val-unseen}}} & \cellcolor{graycell}{\textbf{sanitary}}& \cellcolor{graycell}{bath sheet, bleach cleanser, diaper pack, dishtowel, dustpan and brush, electric toothbrush, incontinence pads, parchment sheet, sanitary pads, soap dish, soap dispenser, sponge, sponge dish, tampons, toothbrush holder, toothbrush pack, towel, washcloth, wipe warmer}\\
        \hline
        & \textbf{cosmetic}& beard color gel, beauty pack, face moisturizer, hair color, hair conditioner, lipstick, mascara, skin care product, skin moisturizer\\
        & \cellcolor{graycell}{\textbf{crockery}}& \cellcolor{graycell}{bowl, cup, dog bowl, drink coaster, mug, stacking cups, tray, tumbler}\\
        & \textbf{electronic accessory}& battery, electronic adapter, electronic cable, graphics card, hard drive, hard drive case, headphones, ink cartridge, keyboard, laptop cover, laptop stand, motherboard, mouse, mouse pad, movie dvd, multiport hub, phone armband case, phone stand, remote control, software cd, tablet holder, tablet stand, usb drive, wireless accessory\\
        & \cellcolor{graycell}{\textbf{fruit}}& \cellcolor{graycell}{apple, banana, lemon, orange, peach, pear, plum, strawberry}\\
        & \textbf{furnishing}& cushion, neck rest, pillow\\
        \multirow{4}{*}{\rotatebox[origin=c]{90}{\textbf{test-unseen}}} & \cellcolor{graycell}{\textbf{stationery}}& \cellcolor{graycell}{book, crayon, file sorter, folder, invitation card, labeling tape, large marker, letter holder, paint bottle set, paint maker, pencil case}\\
        & \textbf{storage}& backpack, bookend, box, canister, carrying case, cube storage box, desk caddy, easter basket, jar, jewelry box, laundry box, lunch bag, lunch box, paper bag, shoe box, snack dispenser, storage bin, waste basket\\
        & \cellcolor{graycell}{\textbf{tool}}& \cellcolor{graycell}{adjustable wrench, anti slip tape, chain, clamp, duct tape, flat screwdriver, hammer, magnifying glass, measuring tape, padlock, phillips screwdriver, power drill, scissors, vinyl tape}\\
        & \textbf{toy}& action figure, android figure, balancing cactus, board game, card game, clay, colored wood blocks, dog chew toy, dollhouse toy, fingerpaint, foam brick, hand bell, jenga, lego duplo, nine hole peg test, nintendo switch, peg and hammer toy, puzzle game, rubiks cube, sidewalk chalk, sorting toy, stuffed toy, toy airplane, toy animal, toy basketball, toy bowling set, toy construction set, toy fishing, toy food, toy furniture set, toy instrument, toy kitchen set, toy tool kit, toy vehicle, video game, whale whistle\\
        \hline
        \end{tabular}
}
\label{table:object_categories}
\end{table}

\clearpage

\section{AMT Human Preferences Dataset}
\label{sec:supp:amt_study} 

In this section, we provide more details on our AMT study interface and perform some analysis on the collected data. Our interface consists of an instructions section and is followed by the main task section. After completing the task, the participants are allowed to submit feedback on the interface and the task. The video at \small{\href{https://www.youtube.com/watch?v=BcHmSzoNBYw}{https://www.youtube.com/watch?v=BcHmSzoNBYw}} walks through our AMT data collection interface. %

\subsection{Participant Instructions}
\label{ssec:supp:amt_ins} 

\begin{figure*}[t!]
    \centering
    \includegraphics[width=\textwidth]{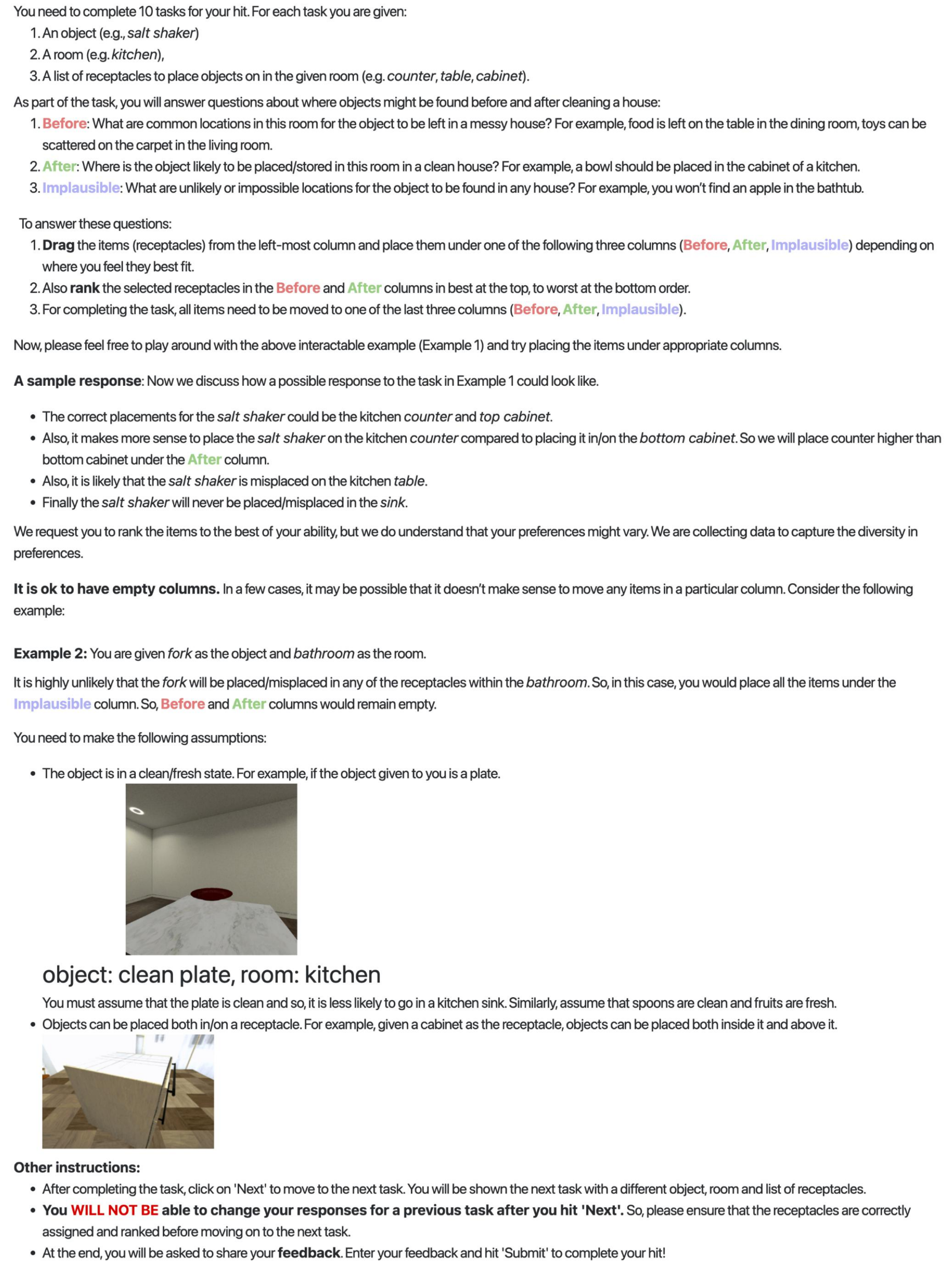}
    \caption{\small{AMT Instructions page describing the task with illustrative examples.}}
    \label{fig:amt_instructions}
\end{figure*}

Before beginning the study, each participant is required to read the instructions section. We show the full set of instructions we used during data collection in Figure~\ref{fig:amt_instructions}. In our instructions, we describe the tasks that need to be performed to successfully complete a HIT (Human Intelligence Task; an AMT term for a unique task instance). As part of a single HIT, the participants are required to complete 10 sub-tasks. For each sub-task, the participant is given an object, a room and a list of receptacles within the given room. The participant is required to classify  these receptacles as \texttt{correct}, \texttt{misplaced} and \texttt{implausible} locations. For the receptacles put into the \texttt{correct} and \texttt{misplaced} bins, the participant is also required to provide a relative ordering between receptacles.

The instructions section includes an interactive example that the participants can use to practice before they work on the actual tasks. As a part of our instructions, we provide multiple examples of valid responses. We ask the participants to assume the object is in its ``base" state (\eg utensils being clean, packaged food being unopened) before making their placement decisions. 

\subsection{Task Interface}
\label{ssec:supp:amt_inter} 
We now describe the task interface in detail. We use the same examples that were used to train the participants.

\begin{figure*}[t!]
    \centering
    \includegraphics[width=\textwidth]{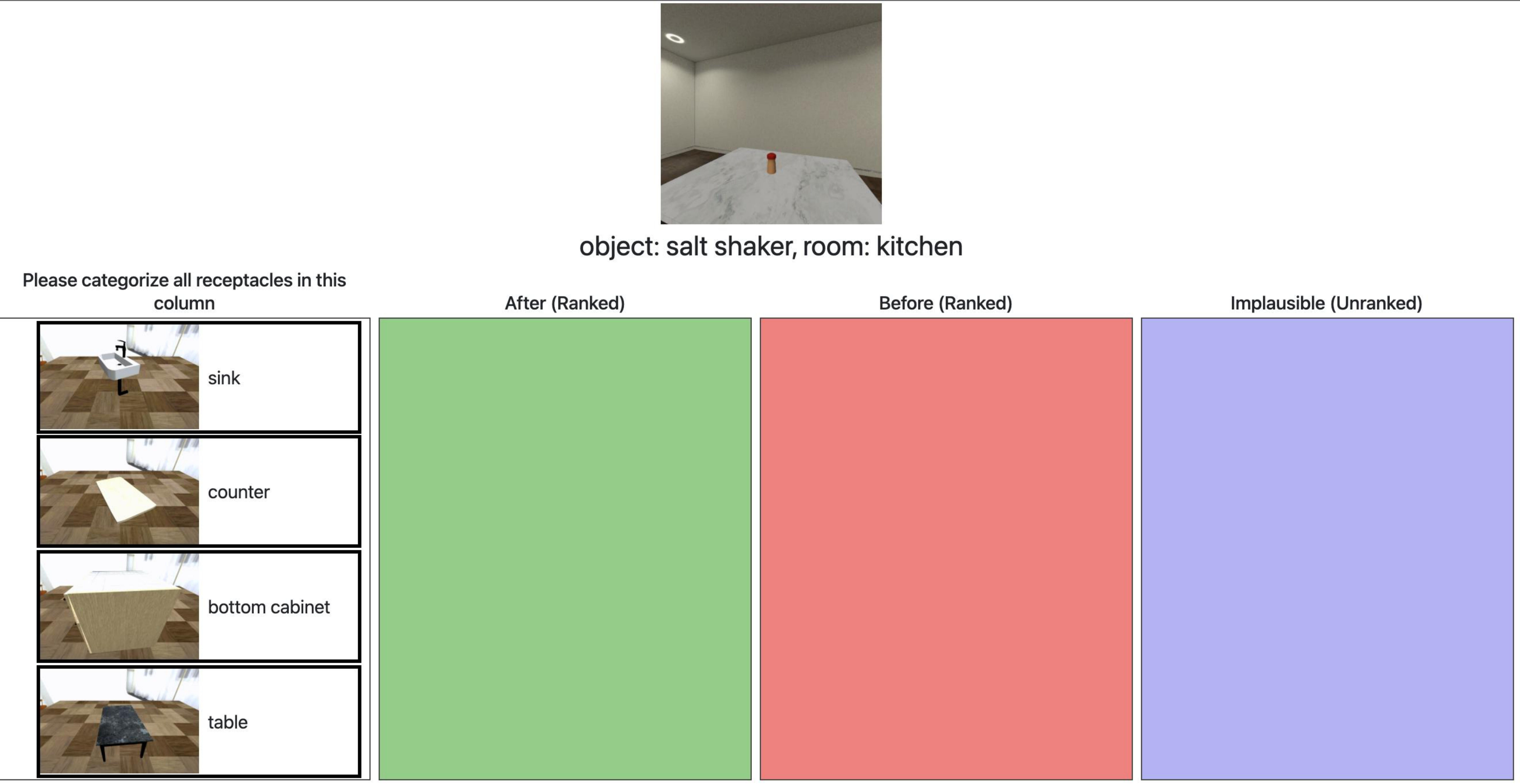}
    \caption{\small{
        AMT starting interface for categorizing and ranking receptacles in the kitchen for a salt shaker.
    }}
    \label{fig:amt_task1}
\end{figure*}

\noindent \textbf{Task Start}: For each sub-task we display an object, a room name and four columns. We show all receptacles to be categorized in the first column, with empty correct and misplaced columns (ranked), and an empty implausible column. The object and receptacles are displayed as rotating animated GIFs. Figure~\ref{fig:amt_task1} shows a screenshot of our task interface at the start of the task. In this example, the receptacles within the kitchen are to be classified as being the correct, misplaced and implausible locations for the alt shaker.

\clearpage

\begin{figure*}[h!]
    \centering
    \includegraphics[width=\textwidth]{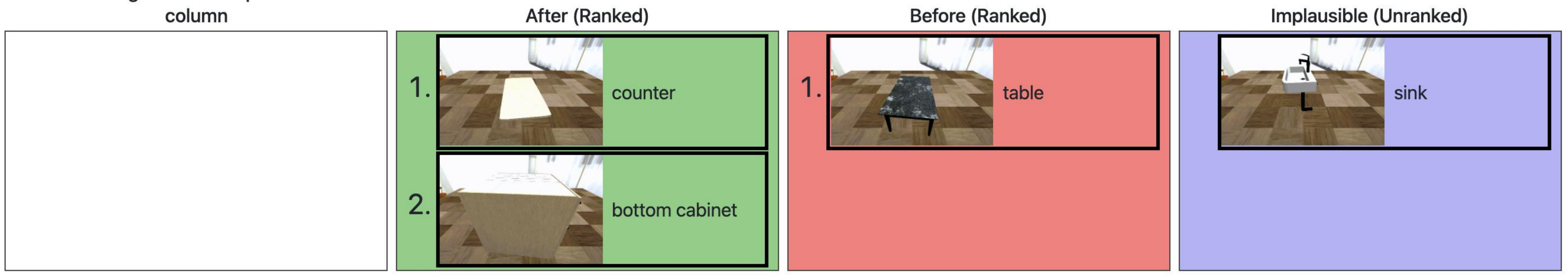}
    \caption{\small{AMT Example 1: A sample response for salt shaker on receptacles in the kitchen provided as an example to the users.}}
    \label{fig:amt_task2}
\end{figure*}

\begin{figure*}[h!]
    \centering
    \includegraphics[width=\textwidth]{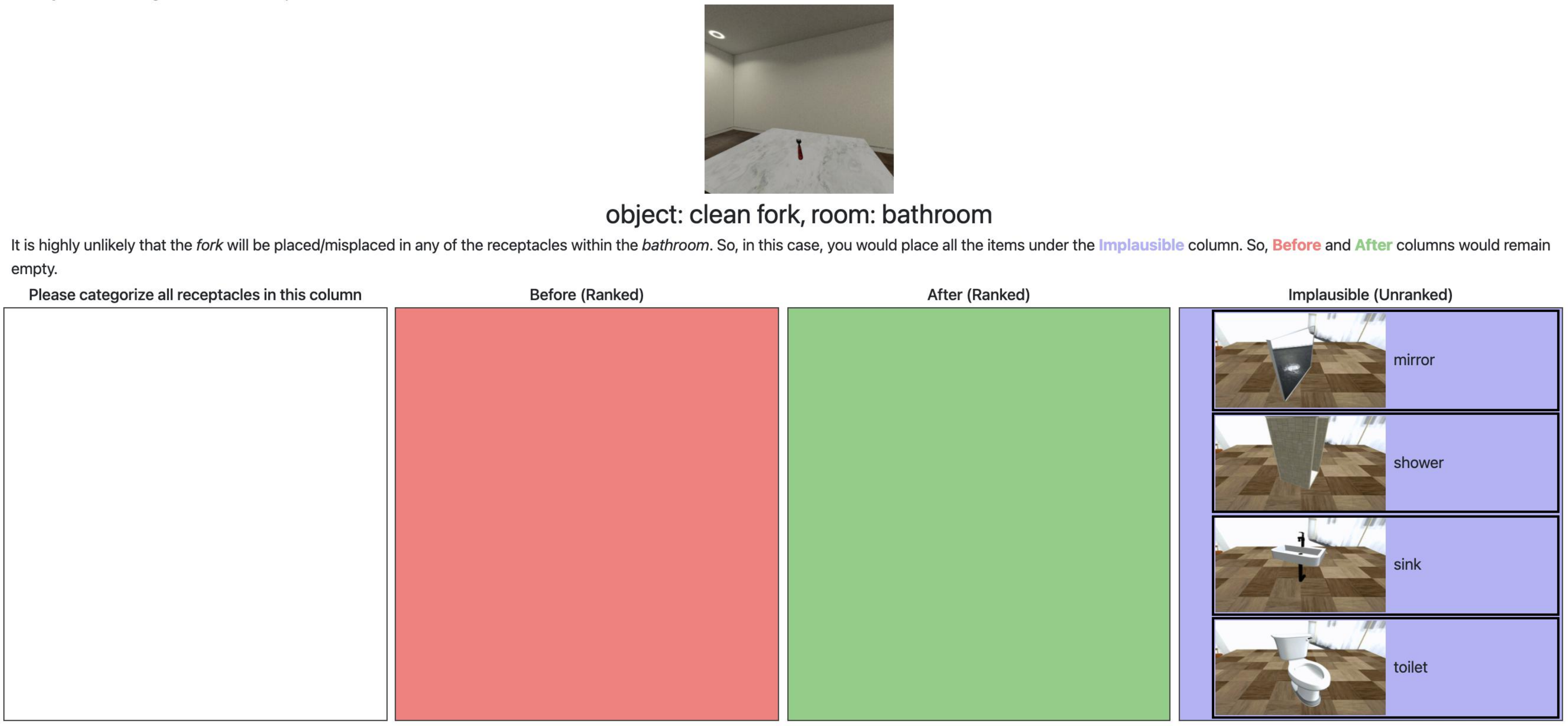}
    \caption{\small{AMT Example 2: A sample response for clean fork on receptacles in the bathroom.}}
    \label{fig:amt_task3}
\end{figure*}

\noindent \textbf{Sample Response \#1}: Figure~\ref{fig:amt_task2} shows a sample response for the task in Figure~\ref{fig:amt_task1}.

\noindent \textbf{Sample Response \#2}: Now consider the example in Figure~\ref{fig:amt_task3}. Here the given object is fork and the given room is bathroom. Since any receptacle within the bathroom is unlikely to be a correct/misplaced location for fork, all receptacles are placed under the Implausible column.

\subsection{Dataset statistics}
We collect 10 annotations for each object-room pair. We consider that a room-receptacle (\eg kitchen-sink) is \textit{selected} as being a correct/misplaced location for a given object (\eg sponge) if at least 6 annotators place the receptacle (\eg sink) under the correct/misplaced column when shown the given object-room pair (\eg sponge-kitchen). Figure~\ref{fig:selected_receptacles_hist} shows a histogram of objects across different numbers of room-receptacles selected as correct or misplaced. We see that fewer room-receptacles are selected as correct placement of objects while most receptacles are selected as incorrect. Additionally, for most objects (\url{~}70\%), annotators selected fewer than 20 receptacles across all rooms as correct. On the other hand, annotators tend to select 10-50 receptacles across all rooms as incorrect placements for most objects. This is also confirmed by Figure~\ref{fig:selected_receptacles_boxplot}. It shows  the distribution of the number of room-receptacles \textit{selected} as being the correct and misplaced locations. 
More receptacles are selected as locations where objects are misplaced compared to receptacles where objects are correctly placed.
\begin{figure}[h!]
  \centering
   \begin{subfigure}[t]{0.485\columnwidth}
    \includegraphics[width=\textwidth,]{figures/selected_placements_hist.pdf}
    \caption{
      \small
       Histogram of objects across different number of room-receptacles selected as correct or misplaced.
    }
    \label{fig:selected_receptacles_hist}
  \end{subfigure} \hfill
  \begin{subfigure}[t]{0.485\columnwidth}
    \includegraphics[width=\textwidth,]{figures/selected_placements_boxplot.pdf}
    \caption{
      \small
      Distribution per high-level category
    }
    \label{fig:selected_receptacles_boxplot}
  \end{subfigure}
  \caption{
    \small
    Number of room-receptacles selected as Correct and Misplaced.
  }
\end{figure}

\clearpage
\section{\tname}
\label{sec:supp:hk} 

\subsection{Episode Generation}
Algorithm~\ref{alg:ep_gen} provides the logic used to generate an episode in \tname. We start with an empty scene \var{S} furnished with receptacles, AMT data \var{D}, objects repository \var{O}. Next, we filter objects by keeping only the ones that have at least one \emph{correct} receptacle in the scene, and remove the others. After initializing an incorrectly placed object, we ensure that the agent is able to rearrange and place it on at least one of the \emph{correct} receptacles. On the other hand, after initializing a correctly placed object, we just ensure that the agent is able to navigate to within grasping distance of it.

\label{sec:supp:episode_generation} 
\begin{algorithm}
    \scriptsize
    \caption{\small \label{alg:ep_gen} Dataset Generation}
    \textbf{import modules:} episode \texttt{E}; human-data \texttt{D}; objects \var{O}, scene \var{S} \\
    \textbf{input variables}: misplaced objects \var{n$_m$}; correct objects \var{n$_c$} \\
    \SetInd{0.2em}{0.5em}
    \Function{build\_episode(\var{E}, \var{D}, \var{O}, \var{S}, \var{n$_m$}, \var{n$_e$}):}{
        \var{\textcolor{gray}{\# initialize and load modules}} \\
        \var{E.init\_empty(), D.load(), S.load(), O.load()} \\
        
        \var{\textcolor{gray}{\# keep only objects that have at least one correct receptacle in the scene}} \\
        \var{objs = S.filter\_objects(O,D)}
        
        \var{\textcolor{gray}{\# insert misplaced  objects}} \\
        \While{len(E.objs) < n$_m$:}{
            \var{\textcolor{gray}{\# sample object to misplace}} \\
            \var{obj = S.sample\_misplaced\_object()} \\
            \var{\textcolor{gray}{\# get corresponding correct and misplace receptacles}} \\
            \var{correct\_recs, misplace\_recs = S.get\_recs(obj)} \\ 
            \var{\textcolor{gray}{\# place object for rearrangement, ensure it is solvable}} \\
            \If{E.place(obj, misplace\_recs) and E.check\_solvable(obj):}{
                \var{E.register(obj)}
            }
        }

        \var{\textcolor{gray}{\# insert correctly placed objects}} \\
        \While{len(E.objs) < n$_m$+n$_c$:}{
            \var{\textcolor{gray}{\# sample object to place correctly}} \\
            \var{obj = S.sample\_placed\_object()} \\
            \var{\textcolor{gray}{\# get correct receptacles only}} \\
            \var{correct\_recs, \_ = S.get\_recs(obj)} \\ 
            \var{\textcolor{gray}{\# place object on correct receptacle, ensure it is graspable}} \\
            \If{E.place(obj, correct\_recs) and E.check\_graspable(obj):}{
                \var{E.register(obj)}
            }
        }
    }
\end{algorithm}

\subsection{Episode statistics}
\label{ssec:supp:rec_use} 

\begin{figure}[h!]
        \label{fig:episode_stats}
        \centering
      \begin{subfigure}[c]{0.6\columnwidth}
      
          \includegraphics[width=\textwidth]{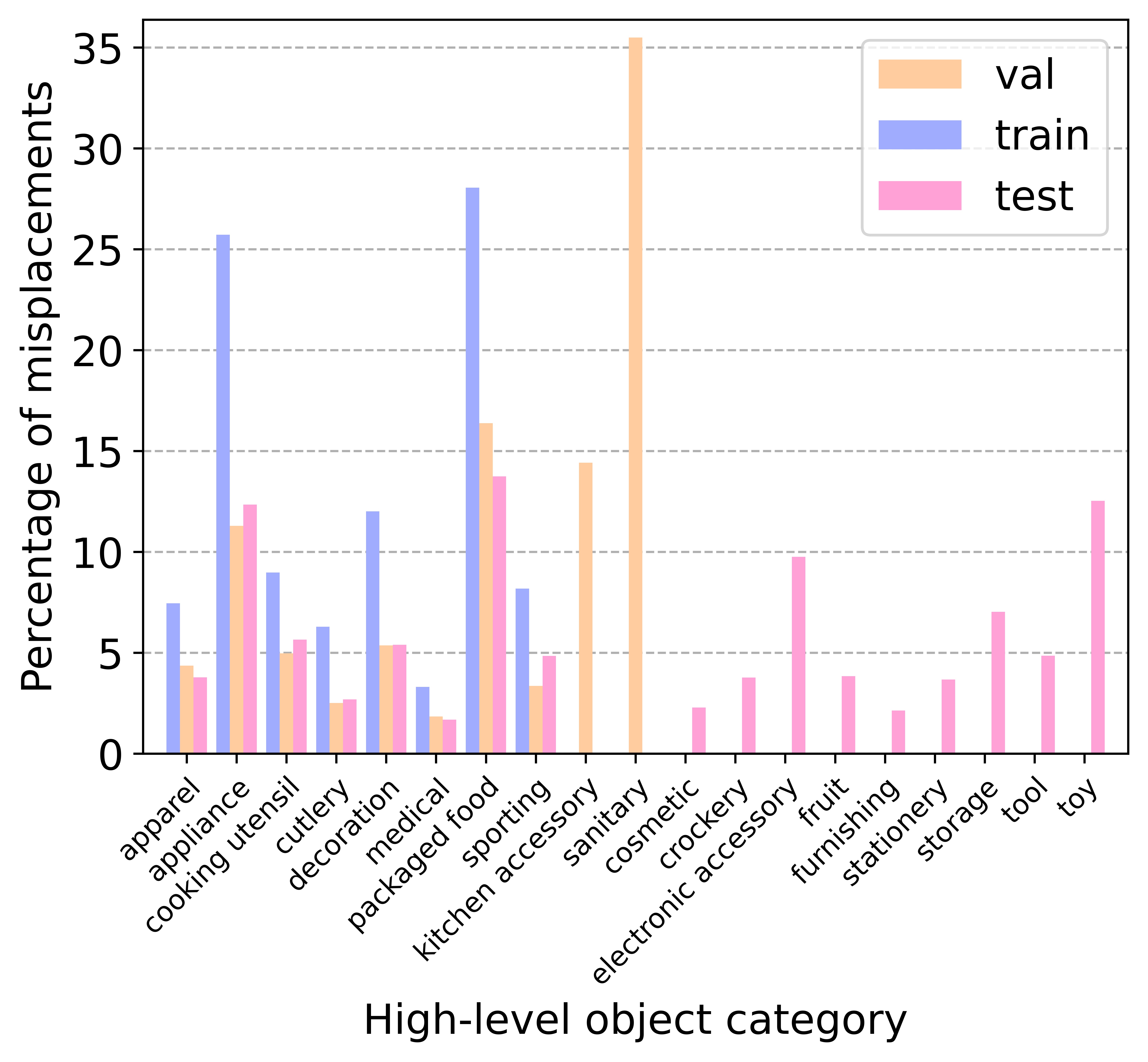}
          \caption{
            \small
            Histogram of misplaced objects in episodes across different high-level object categories
          }
          \label{fig:misplacements_per_high_level_cat}
      \end{subfigure}\\
      \begin{subfigure}[t]{0.45\columnwidth}
          \centering
          \includegraphics[width=0.95\textwidth]{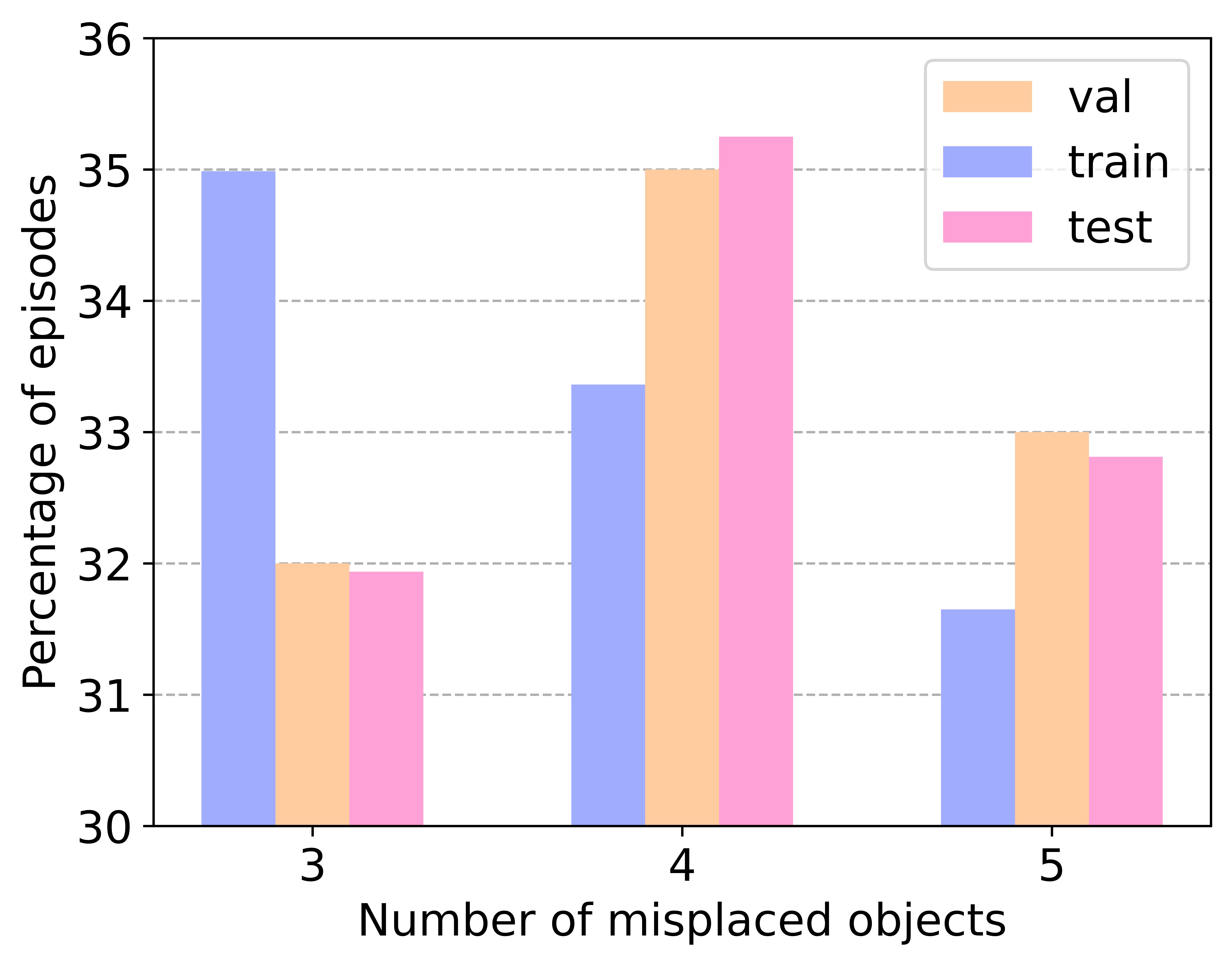}
          \captionsetup{width=0.95\textwidth}
          \caption{
            \small
            Histogram showing percentage of train, val and test episodes with given number of misplaced objects
          }
          \label{fig:misplacements_per_episode}
      \end{subfigure}
      \begin{subfigure}[t]{0.45\columnwidth}
          \centering
          \includegraphics[width=0.95\textwidth]{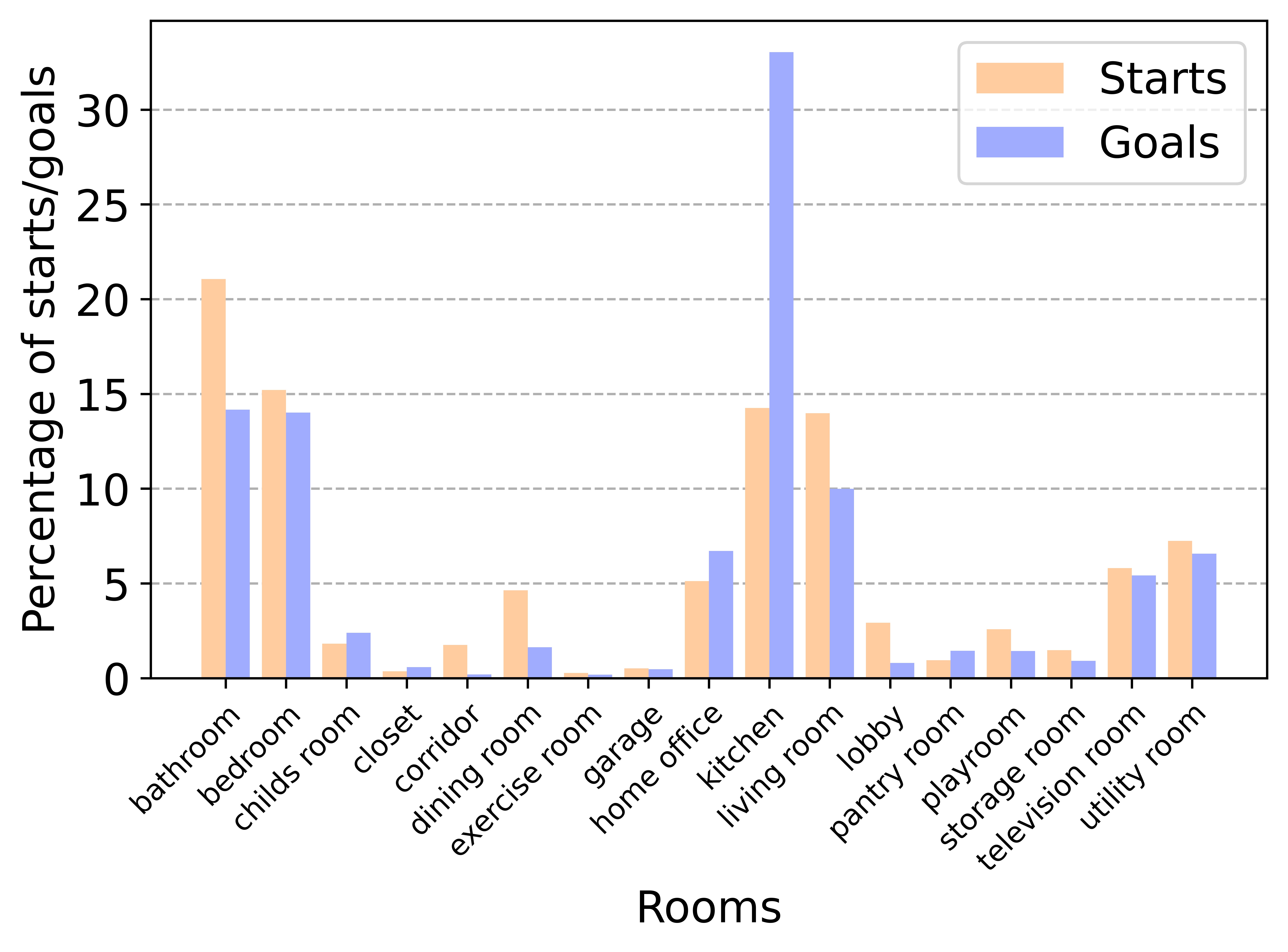}
          \captionsetup{width=0.95\textwidth}
          \caption{
            \small
            Histogram showing percentage of start and goal positions in each room
          }
          \label{fig:start_goal_across_rooms}
      \end{subfigure}
      \caption{
        \small
        \textbf{Episode Statistics.} Analysis on misplaced objects in episodes and their start and goal positions
      }
\end{figure}

    \begin{figure}[h!]
      \begin{subfigure}[t]{0.45\columnwidth}
          \centering
          \includegraphics[width=0.9\textwidth]{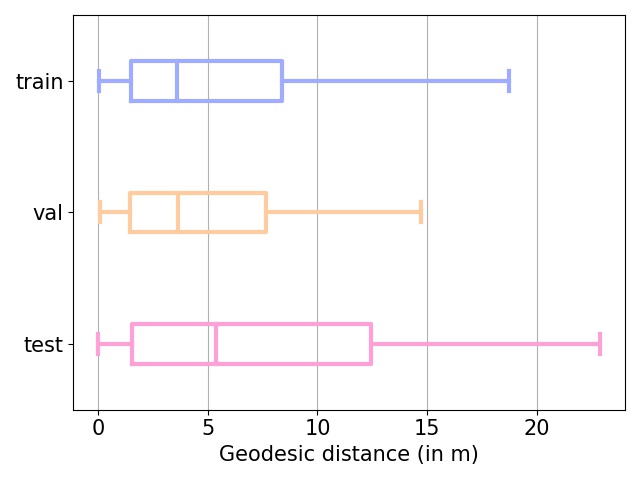}
          \caption{\small{Start to every goal}}
          \label{fig:distance_to_all_goals}
      \end{subfigure}
      \begin{subfigure}[t]{0.45\columnwidth}
          \centering
          \includegraphics[width=0.9\textwidth]{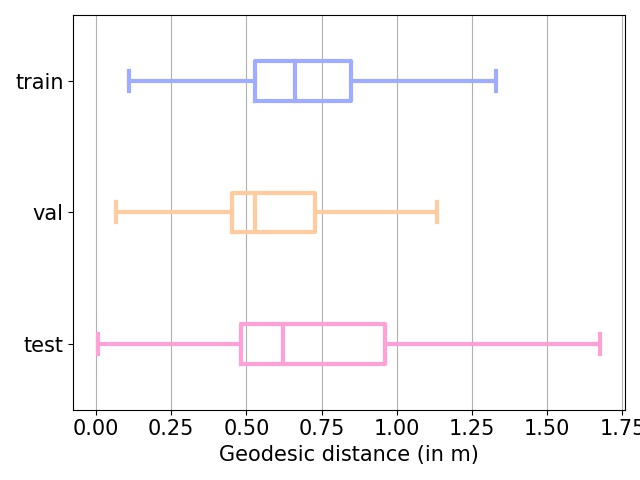}
          \caption{\small{Start to closest goal}}
          \label{fig:distance_to_closest_goal}
      \end{subfigure}
      \label{fig:geodesic}
      \caption{
        \small{Distribution of geodesic distance from start receptacle to (a) every goal (b) closest goal.}
      }
    \end{figure}
    
    We analyze the generated train, val and test episodes. The val and test episodes include high-level categories already seen in train episodes as well as a few novel high-level categories (Figure~\ref{fig:misplacements_per_high_level_cat}). Each episode in the train, val and test splits has $3-5$ misplaced objects. Our val and test episodes have slightly higher percentages of episodes with 4 or 5 misplaced objects compared to train episodes (Figure~\ref{fig:misplacements_per_episode}). A large fraction of the misplaced objects in our episodes start in a bathroom, bedroom, kitchen or living room. A large number of goal receptacles for the misplaced objects are located in the kitchen~\ref{fig:start_goal_across_rooms}. This is expected since a large number of misplaced objects in a household usually are food or cooking-related (see Figure~\ref{fig:misplacements_per_high_level_cat}), and kitchens usually have a large number of receptacles.
    
\noindent \textbf{Object-Receptacle Distances:} Next, we visualize the distribution of geodesic distances from object to correct receptacles across all misplaced objects in all episodes. The median distance in our test episodes is 5.36m (Figure~\ref{fig:distance_to_all_goals}) and the median distance to the closest correct receptacle (out of the 3-5 mispalced) in the test episodes is 0.62m (Figure~\ref{fig:distance_to_closest_goal}).

\subsection{Formal definitions of metrics}
\label{sec:supp:eval} 

In \Cref{sec:task:eval}, we informally described our evaluation metrics for \tname. Here, we formally define the metrics for which more rigorous explanations are required.

For a given scene, $\gR$ and $\gO$ are the set of all receptacles and objects respectively. Given an object $o \in \gO$, let $c_{or}$, $m_{or}$ respectively be the ratio of annotators who placed receptacle $r \in \gR$ in \var{correct} and \var{misplaced} bins respectively. We call an object \emph{correctly placed} if $c_{or} > 0.5$, and \emph{misplaced} if $m_{or} > 0.5$, where both cannot be simultaneously true. We use:
\begin{compitemize}
\item $\gO_{m}$ for the set of objects which were \emph{initially misplaced} in the episode.
\item $\gO_{i}$ for the set of objects which were \emph{interacted} with by the agent during the episode.
\item $\gO_{mi}$ ($\gO_{i} \cup \gO_{m}$) for the set of objects \emph{initially misplaced} or \emph{interacted} with by the agent during the episode.
\end{compitemize}

Finally, we define the final placement of the object $o$ at the end of the episode via a mapping function $\Phi: \gO \rightarrow \gR$. The receptacle on which an object $o \in \gO$ is placed at the end of the episode is given by $\Phi(o)$

Given the relative change in placement of objects between the start and end states of the episode ($\gS_1$ vs $\gS_{T}$), we can formally write the rearrangement metrics as:

\begin{compenumerate}
    \item \textbf{Episode Success (\var{ES})}: Strict binary (\emph{all} or \emph{none}) metric that is one if and only if all objects are correctly placed, \var{ES}$=\prod_{o \in \gO} \mathbbm{1}[{c_{o, \Phi(o)} > 0.5]}$.
    \item \textbf{Object Success (\var{OS})}: Fraction of the objects which were \emph{initially misplaced} or \emph{interacted} with by the agent placed correctly at end of the episode, \var{OS}$=\sum_{o \in \gO_{mi}} \mathbbm{1}[{c_{o, \Phi(o)} > 0.5]}/|\gO_{mi}|$.
  \item \textbf{Soft Object Success (\var{SOS})}: The ratio of reviewers that agree that every object \emph{interacted} with or \emph{initially misplaced} is placed correctly averaged across all rearranged objects, \var{SOS}$=\sum_{o \in \gO_{mi}} c_{o,\Phi(o)}/|\gO_{mi}|$.
  This metric is more lenient because it will be a non-zero number even if just one annotator thought the mapping $(o, \phi(o))$ is correct.
  \item \textbf{Rearrange Quality (\var{RQ})}: The normalized ranking in $(0, 1]$ (via mean reciprocal rank ~\cite{mrr}) of the receptacle on which an object is placed, ranked among all correct receptacles of an object, if the object was correctly placed, 0 otherwise, averaged across all \emph{initially misplaced} or \emph{interacted} objects. \var{RQ}$=\sum_{o \in \gO_{mi}} \mathbbm{1}[c_{o, \Phi(o)} > 0.5] mrr_{c_{o, \Phi(o)}}.$ Intuitively, RQ will score higher those rearrangements that have a high overall rank in the human preferences dataset.
\end{compenumerate}

To formally define Pick and Place Efficiency (\var{PPE}), one of our exploration metrics, we need a few extra definitions.

We define $N: \gO_i \rightarrow \{1, 2, \cdots\}$ to be a function mapping an object $o \in \gO_{i}$ to the number of times it was \emph{picked} or\emph{ placed} by the agent. We similarly define $N_{min}: \gO_i \rightarrow \{0, 2\}$ to be the minimum number of picks and places to place an object $o \in \gO_{i}$ in a correct receptacle: it is 2 when $o \in \gO_{m}$ and 0 otherwise.

\textbf{Pick and Place Efficiency (\var{PPE})}: The minimum number of interactions needed to rearrange an object divided by the number of interactions the agent actually took to rearrange it if the object was placed in the correct receptacle by the agent at the end of the episode, and 0 if the object was in the incorrect receptacle at the end of the episode, averaged across all objects the agent \emph{interacted} with. \var{PPE} $= \sum_{o \in \gO_{i}} \mathbbm{1}[c_{o, \phi(o)} > 0.5] \frac{N(o)}{N_{min}(o))} / |\gO_{i}|$

\section{Agent}
\label{sec:supp:navpp} 

We expand on low-level modules used in the agent for navigation and pick-place. %

\noindent \textbf{\texttt{Navigation (N): }} Indoor navigation between two points (aka PointNav) is a well-studied problem both in embodied AI~\cite{Wijmans2020DDPPOLN, ZhaoICCV2021, Ye2020AuxiliaryTS} and classical robotics~\cite{Chan2018Robust2I, 8968455, indoor-robot}. Our navigation module takes as input the allocentric map and a goal position (object, receptacle, or frontier), and executes a sequence of low-level base control actions to reach the goal.\smallskip

\noindent \textbf{\texttt{Pick-Place (P): }} Recall from Section~\ref{sssec:task} that to interact with an object, the agent invokes a discrete action that casts a ray, and if it intersects an object or receptacle within 1.5m of the agent, it picks or places the object.
Our hierarchical baseline picks and places objects via the instance ID of an object or receptacle currently in the view of the agent.
The agent then orients itself to face the desired instance ID via look up/down and turn left/right actions.
Once the desired instance ID is within the agent's view, the agent calls the ray-cast interaction action.
The Pick-Place module fails if the agent is unable to view the object/receptacle of interest or navigate to a place within interaction distance.   
However, we ensure all episodes are solvable by an oracle agent, so this does not occur in the episodes on which we run our hierarchical baseline.
The Pick-Place module can also fail to place an object on a receptacle if sufficient space is not available on the receptacle.

\section{Approach}
\label{sec:supp:approach} 
\subsection{LLM Ranking Module}
In Table~\ref{tab:orm_details}, we provide the hyperparameters that we use to train the \texttt{OR} and \texttt{ORR} modules using the contrastive matching (\texttt{CM}) strategy. 
Each method trained using \texttt{CM} is trained on a single GPU for 1000 epochs and we choose the training checkpoint that gives the best mAP score (evaluated as in Section~\ref{sec:exp:llm}) on the validation set. In the case of \texttt{RoBERTa+CM}, we use the pretrained roberta-base model and average the last-layer hidden state at all positions (including the CLS token) to obtain the text embeddings.
\begin{table}[h!]
    \centering
    \caption{\small{Hyperparameter choices for training the \texttt{CM} modules}}
    \setlength{\tabcolsep}{10pt}
    \begin{tabular}{c l c}
        \toprule
		\textbf{\#} & \textbf{Hyperparameter} & \textbf{Value} \\
		\midrule
        1 & Embedding size & 768 (RoBERTa) / 300 (GloVe) \\
        2 & MLP hidden dimension & 512 \\ 
        3 & MLP out dimension & 512 \\
        4 & MLP hidden layers & 2 \\
        5 & Batch size & 64 \\
        6 & Optimizer & Adam \\
        7 & Learning rate & 0.01 \\
        8 & Weight decay & 0.2 \\
        \bottomrule
    \end{tabular}
    \label{tab:orm_details}
\end{table}

\section{Additional Experiments}
\label{sec:supp:add_exps} 

\subsection{Exploration Strategies}
\label{ssec:supp:expl_strats}

\begin{table*}[t]
	\setlength{\tabcolsep}{3pt}
	\caption{\small{Evaluation of exploration strategy on \var{val} split. \texttt{RND}: Random, \texttt{FWR}: Forward-Right, \texttt{FRT}: Frontier}}
	\begin{center}
		\begin{tabular}{c l cccc}
			\toprule

			\scriptsize \texttt{\#} & \scriptsize Strategy & \scriptsize\textbf{\texttt{OS}}~$\uparrow$ & \scriptsize\textbf{\texttt{MC}}~$\uparrow$ &  \scriptsize\textbf{\texttt{OC}}~$\uparrow$ & \scriptsize\textbf{\texttt{PDE}}~$\uparrow$ \\
			\midrule
			
           \scriptsize \texttt{1} & \texttt{RND} & 0.12 \confint{0.01} & 43 \confint{1} & 0.40 \confint{0.02} & 0.22 \confint{0.02} \\
            \scriptsize \texttt{2} & \texttt{FWR} & 0.11 \confint{0.01} & 38 \confint{1} & 0.34 \confint{0.02} & 0.20 \confint{0.02} \\
            \scriptsize \texttt{3} & \texttt{FRT} & 0.26 \confint{0.01} & 86 \confint{2} & 0.76 \confint{0.02} & 0.33 \confint{0.02} \\

            \bottomrule
		\end{tabular}
	\end{center}
	\label{tab:expl:strat}
\end{table*}

In \Cref{sec:methods}, we discussed the \var{Explore} module that used frontier exploration (\var{FRT}). We evaluate 2 additional simple exploration strategies for a total of the following 3 strategies:

\begin{compitemize}
    \itemsep0em 
    \item \texttt{frontier}: Using the egocentric map we iteratively visit unexplored frontiers, frontiers are defined as the edges between known and unknown space. We keep our implementation details same as those used in~\cite{ramakrishnan2020exploration}.
    \item \texttt{random}: Executes a random action in the navigator. 
    \item \texttt{forward-right}: Executes the forward action until a collision occurs, then turns right. 
\end{compitemize}

As we expect, from \Cref{tab:expl:strat} we see that \var{FRT} outperforms \var{RND} and \var{FWD} in \var{OS}, exploration and efficiency metrics.

\subsection{Planner Ablations}
\label{ssec:supp:plan_abl}

\textbf{Rearrangement Ordering}: %
In \Cref{sec:methods}, when discussing the \var{Rearrange} submodule, we mentioned 3 key decisions in the submodule. One of them was the order in which misplaced objects are rearranged. In this section, we evaluate the following 4 ordering schemes:

\begin{table*}[t]
	\setlength{\tabcolsep}{3pt}
	\caption{
    \small{Evaluation of rearrangement ordering on \var{val} split. \texttt{DIS}: DIScovery order, \texttt{SCG}: Score Gain, \texttt{A-O}: Agent-Object distance, \texttt{O-R}: Object-Receptacle distance}
    }.
	\begin{center}
		\begin{tabular}{c l cc}
			\toprule
			\scriptsize \texttt{\#} & \scriptsize Order & \scriptsize\textbf{\texttt{OS}}~$\uparrow$ & \scriptsize\textbf{\texttt{PDE}}~$\uparrow$ \\
			\midrule
			
            \scriptsize \texttt{1} & \texttt{DIS} & 0.27 \confint{0.01} & 0.35 \confint{0.02} \\
            \scriptsize \texttt{2} & \texttt{SCG} & 0.26 \confint{0.01} & 0.34 \confint{0.02} \\
            \scriptsize \texttt{3} & \texttt{A-O} & 0.25 \confint{0.01} & 0.32 \confint{0.02} \\
            \scriptsize \texttt{4} & \texttt{O-R} & 0.25 \confint{0.01} & 0.32 \confint{0.02} \\

            \bottomrule
		\end{tabular}
	\end{center}
	
	\label{tab:rearrange_strat}
\end{table*}

\begin{compitemize}
    \itemsep0em 
    \item \var{score-diff}: We sort rearrangements in decreasing order of score difference between the current receptacle and best one.   
    \item \var{obj-dist}: We sort rearrangements by the geodesic distance from agent to the object. 
    \item \var{rearrange-dist}: We sort rearrangements by the geodesic distance required to execute the rearrangment.
    \item \var{disc-time}: We sort rearrangements by the time of discovery object. 
\end{compitemize}

In \Cref{tab:rearrange_strat}, we see that the \var{DIS} rearrangement ordering performs slightly better than the other orderings. We choose this ordering to run our main experiments.

\noindent \textbf{Exploration Steps}: %
One of the challenges in \tname is balancing the exploration-exploitation trade-off; the agent must explore to find misplaced objects or suitable receptacles, but also must exploit its existing knowledge of where objects belong.
The exploration module in our hierarchical baseline has an adjustable parameter \var{n}$_e$ that controls the number of steps at the beginning of the episode used for exploration.
This parameter thus controls how long the agent spends exploring versus rearranging objects according to a plan.

We find that fewer exploration steps is more effective. 
If the agent spends too long exploring, then it will not have enough time to rearrange objects before the end of the episode.
\eg when \var{n}$_e=512$, our Object Coverage (\var{OC}) is 80\%, which is 4 points ahead of the next best \var{n}$_e$. However, its Object Success (\var{OS}) is the worst among the variants of \var{n}$_e$ we evaluated.
We found the best number of exploration steps to be \var{n}$_e=16$, achieving higher performance in terms of object success (\var{OS}) than all \var{n}$_e<16$ and \var{n}$_e>16$.

\section{More Qualitative Analysis}
\label{sec:supp:qual_analysis} 

\begin{figure*}[t!]
    \centering
    \includegraphics[width=1\textwidth]{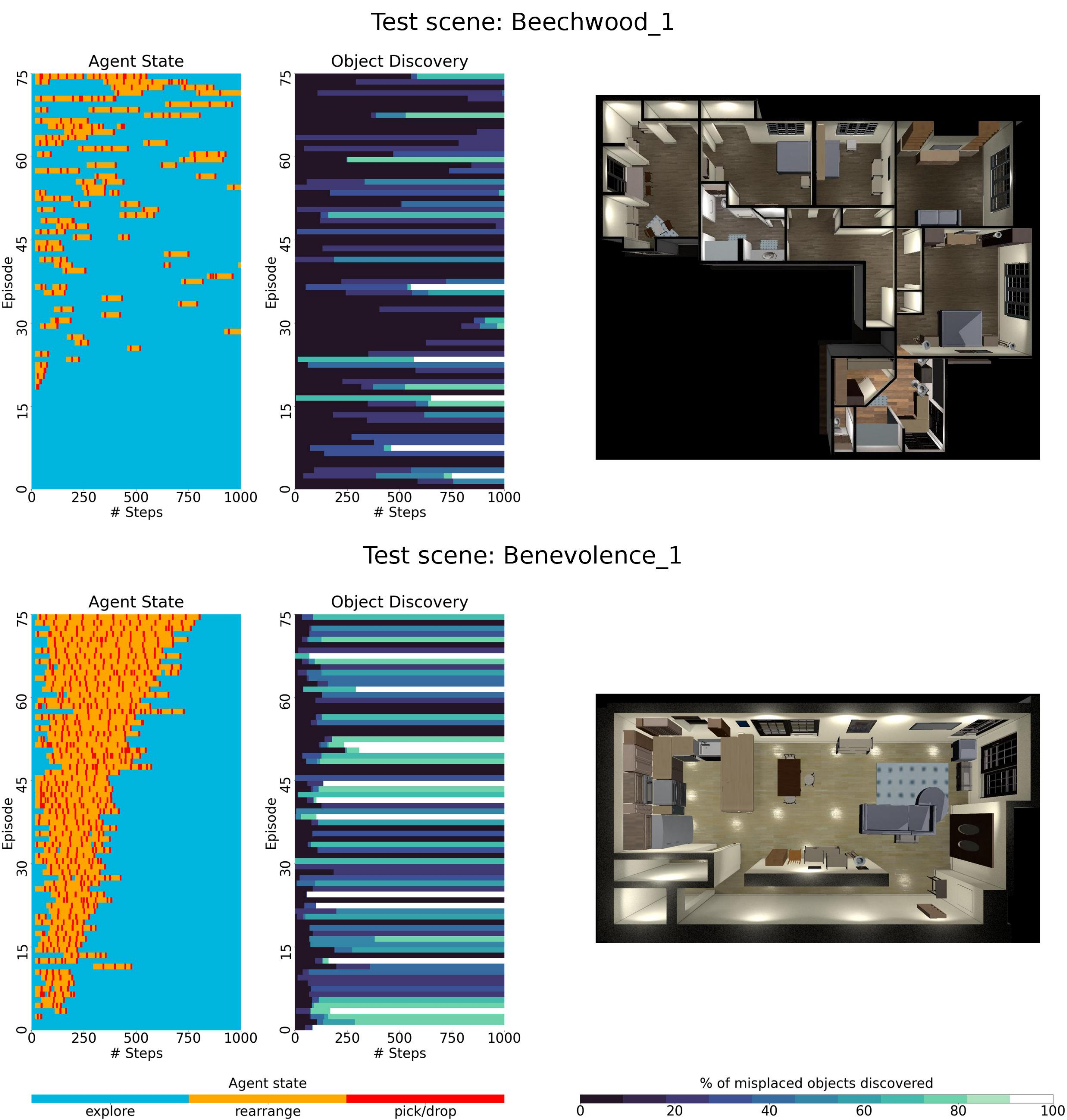}
    \caption{
        \small
        Left column: visually depicting agent's progress on 75 randomly-sampled episodes from two test scenes, \var{beechwood\_1} and \var{benevolence\_1}. Right column: corresponding test scene layouts.
  }
    \label{fig:gantt_2a_layouts}
\end{figure*}

\subsection{Agent states and scene layouts}
\label{ssec:supp:topdown}

\begin{figure*}[ht!]
    \centering
    \includegraphics[width=1\textwidth]{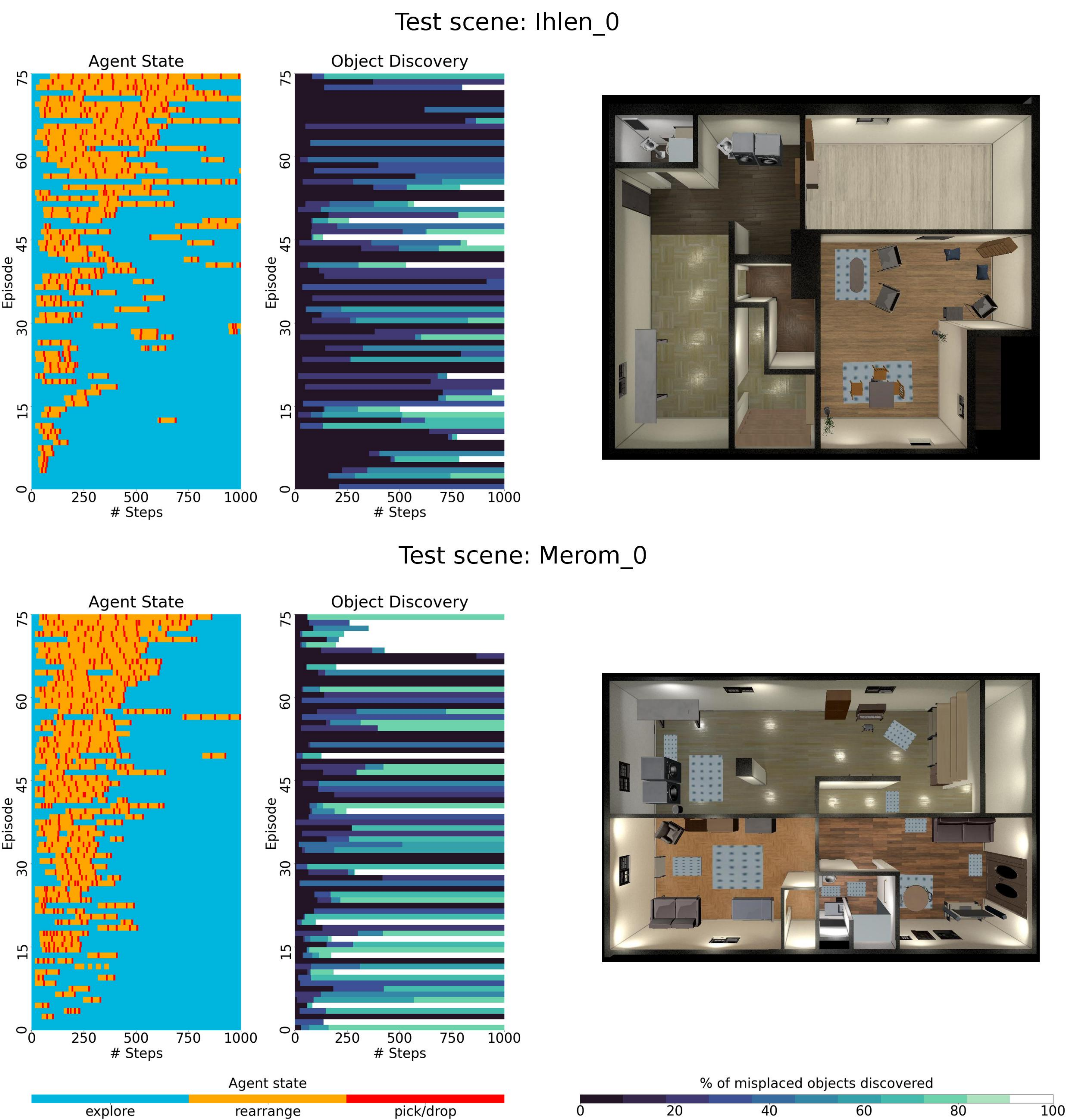}
    \caption{
        \small
        Left column: visually depicting agent's progress on 75 randomly-sampled episodes from two test scenes, \var{ihlen\_0} and \var{merom\_0}. Right column: corresponding test scene layouts. 
  }
    \label{fig:gantt_2b_layouts}
\end{figure*}

\Cref{fig:gantt_2a_layouts} and \Cref{fig:gantt_2b_layouts} contain similar plots to the ones in \Cref{fig:gantt} that were discussed in \Cref{sec:exp:qual_results}.
In particular, we notice that the layout of scene \var{Beechwood\_1} is significantly more complex than that of \var{Benevolence\_1}, which is the cause of the difference between their object discovery plots as discussed in \Cref{sec:exp:qual_results}.
\pagebreak
\section{Egocentric rearrangement video}

We attach an egocentric video (\small{\href{https://www.youtube.com/watch?v=XccBpQNGN1Q}{https://www.youtube.com/watch?v=XccBpQNGN1Q}}) of the agent successfully rearranging all misplaced objects in an episode.
The 3 overlays on the left are, from top to bottom: the depth sensor, instance ID mask with semantic information, and the allocentric top-down occupancy map used by the \var{Mapping} module (see \Cref{sec:methods}).
We also include text logs at the bottom left, showing the object the agent is currently holding, the position and name of the object/receptacle it is navigating towards, the action taken at each step, and whether it is exploring, navigating (rearranging) or picking/placing.

The scene contains 4 misplaced objects: an Easter basket in the utility room table, an electronic adapter and a padlock on the dryer, and a toy vehicle on the sofa. The agent explores until 0:15. It then rearranges the Easter basket, the adapter and the padlock by moving them to a shelf. It completes this rearrangement phase at 1:41, after which it goes back to exploring until 2:07. It then moves the toy vehicle object to a nearby shelf, after which it explores for the remainder of the episode.
\section{Ranking module analysis}
For the main results in the paper (Table~\ref{tab:orm} and Table~\ref{tab:main:unseen}), we used \texttt{RoBERTa+CM} as the scoring function. In this section, we analyze the design choices and the performance of our current ranking module.
\subsection{Ablations}
\begin{table*}[h!]
    
	\setlength{\tabcolsep}{2pt}
	\caption{\small{\textbf{Comparison of features.} \texttt{ORR} and \texttt{OR} results on using different features as text embeddings}}
	\begin{center}

		\begin{tabular}{ll ccc ccc}
			\toprule
            &  & \multicolumn{3}{c}{\textbf{\var{ORR}}} & \multicolumn{3}{c}{\textbf{\var{OR}}} %
            \\
			\cmidrule(lr){3-5} \cmidrule(lr){6-8} 
            \texttt{\#} & \textbf{\var{Features}} & \var{train} &  \var{val-u} &  \var{test-u} & \var{train} & \var{val-u} & \var{test-u} %
            \\
            \midrule
            \texttt{1} & \texttt{CLS} & 0.80 & 0.79 & 0.79 & 0.72 & 0.61 & 0.66
            \\ 
            \texttt{2} & \texttt{Avg-all-exclude-CLS} & 0.82 & 0.79 & 0.80 & 1.0 & 0.61 & 0.66 
            \\ 
            \texttt{3} & \texttt{\textbf{Avg-all}} &  0.81 & 0.79 & 0.81 & 1.0 & 0.65 & 0.65 
            \\ 
            \bottomrule
		\end{tabular}
	\end{center}
	
	\label{tab:orm_features}
\end{table*}

In Table~\ref{tab:orm_features}, we analyze the effect of using different features as the language model text embedding. Our results in the paper use features that are globally averaged over all token positions of the language model (\texttt{Avg-all}). We perform experiments using the features at CLS token (\texttt{CLS}) and using features averaged at all positions except CLS token (\texttt{Avg-all-exclude-CLS}). While the \texttt{Avg-all-exclude-CLS} features perform  close to \texttt{Avg-all} features, using \texttt{CLS} features results in poor performance on seen categories for \texttt{OR} task.

\begin{table*}[h!]
	\setlength{\tabcolsep}{2pt}
	\caption{\small{\textbf{Comparison of language models.} \texttt{ORR} and \texttt{OR} results with different language models}}
	\begin{center}

		\begin{tabular}{ll c ccc ccc}
			\toprule
            &  & & \multicolumn{3}{c}{\textbf{\var{ORR}}} & \multicolumn{3}{c}{\textbf{\var{OR}}} %
            \\
			\cmidrule(lr){4-6} \cmidrule(lr){7-9} 
            \texttt{\#} & \textbf{\var{Method}} & \textbf{\# LLM params.} & \var{train} &  \var{val-u} &  \var{test-u} & \var{train} & \var{val-u} & \var{test-u} %
            \\
            \midrule
            \texttt{1} & \texttt{\textbf{RoBERTa-base+CM}} & 125M & 0.81 & 0.79 & 0.81 & 1.0 & 0.65 & 0.65
            \\ 
            \texttt{2} & \texttt{GPT2+CM} & 117M & 0.84 & 0.79 & 0.83 & 0.92 & 0.62 & 0.64 
            \\ 
            \texttt{3} & \texttt{T5-base+CM}& 220M &  0.85 & 0.82 & 0.84 & 0.95 & 0.69 & 0.68
            \\ 
            \bottomrule
		\end{tabular}
	\end{center}
	
	\label{tab:orm_lang_model}
\end{table*}

Next, we replace the embeddings from RoBERTa-base model with embeddings from GPT-2 and T5-base language models. Note that we use \texttt{Avg-all} features for all language models. We find that using T5-base model results in superior performance on both \texttt{OR} and \texttt{ORR} tasks (Table~\ref{tab:orm_lang_model}). The T5-base model has nearly double the number of parameters in RoBERTa-base model. We compare to T5-base model because the next smaller model, T5-small has 60 million parameters (half the number of parameters in RoBERTa-base).

\subsection{High-level category-wise performance}
\begin{figure}
    \centering
    \includegraphics[width=0.75\textwidth]{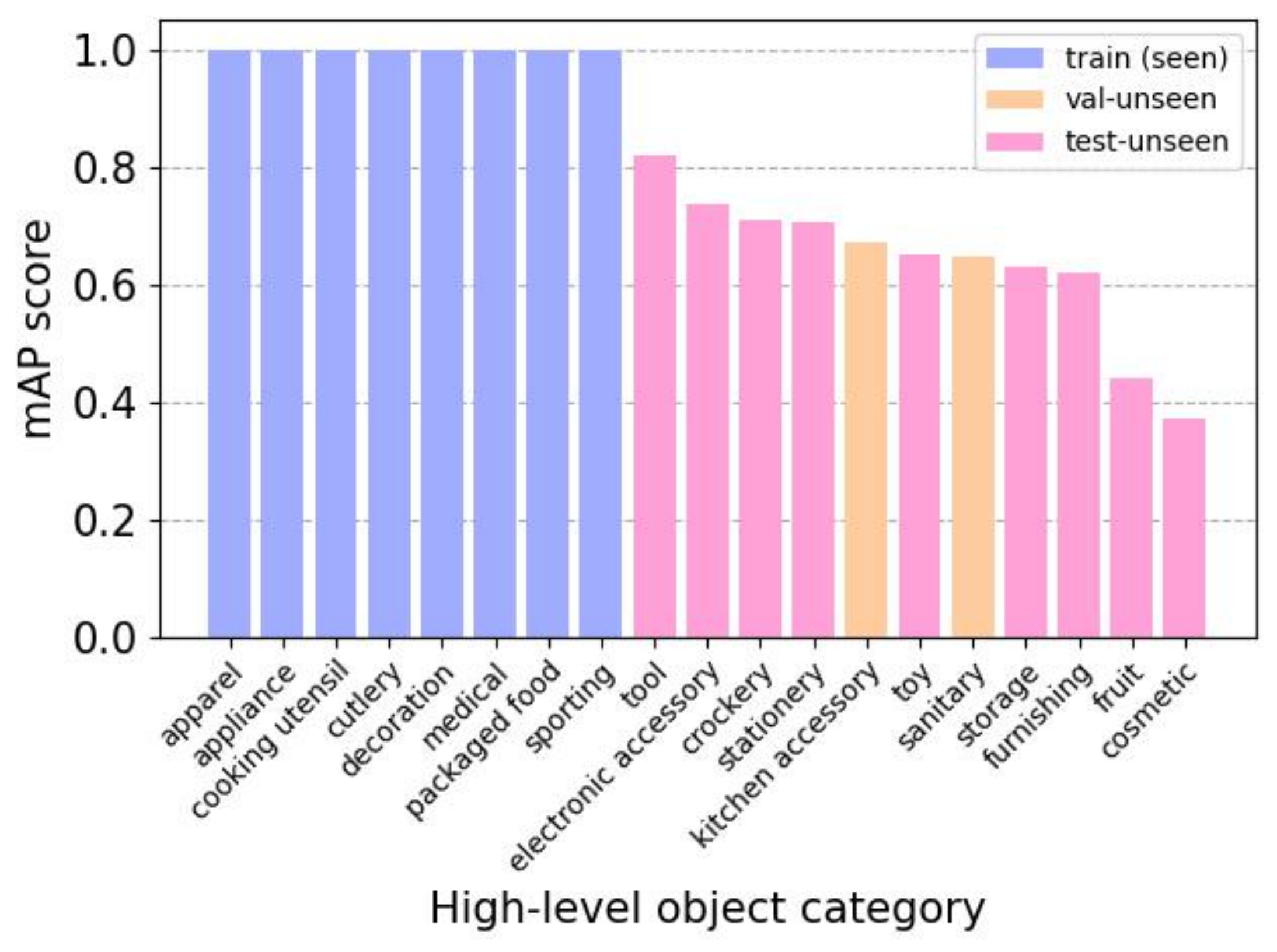}
    \caption{\small{\texttt{OR} performance of \texttt{RoBERTa + CM} across different high-level categories}}
    \label{fig:or_per_high_level}
\end{figure}

\begin{figure}
    \centering
    \includegraphics[width=0.75\textwidth]{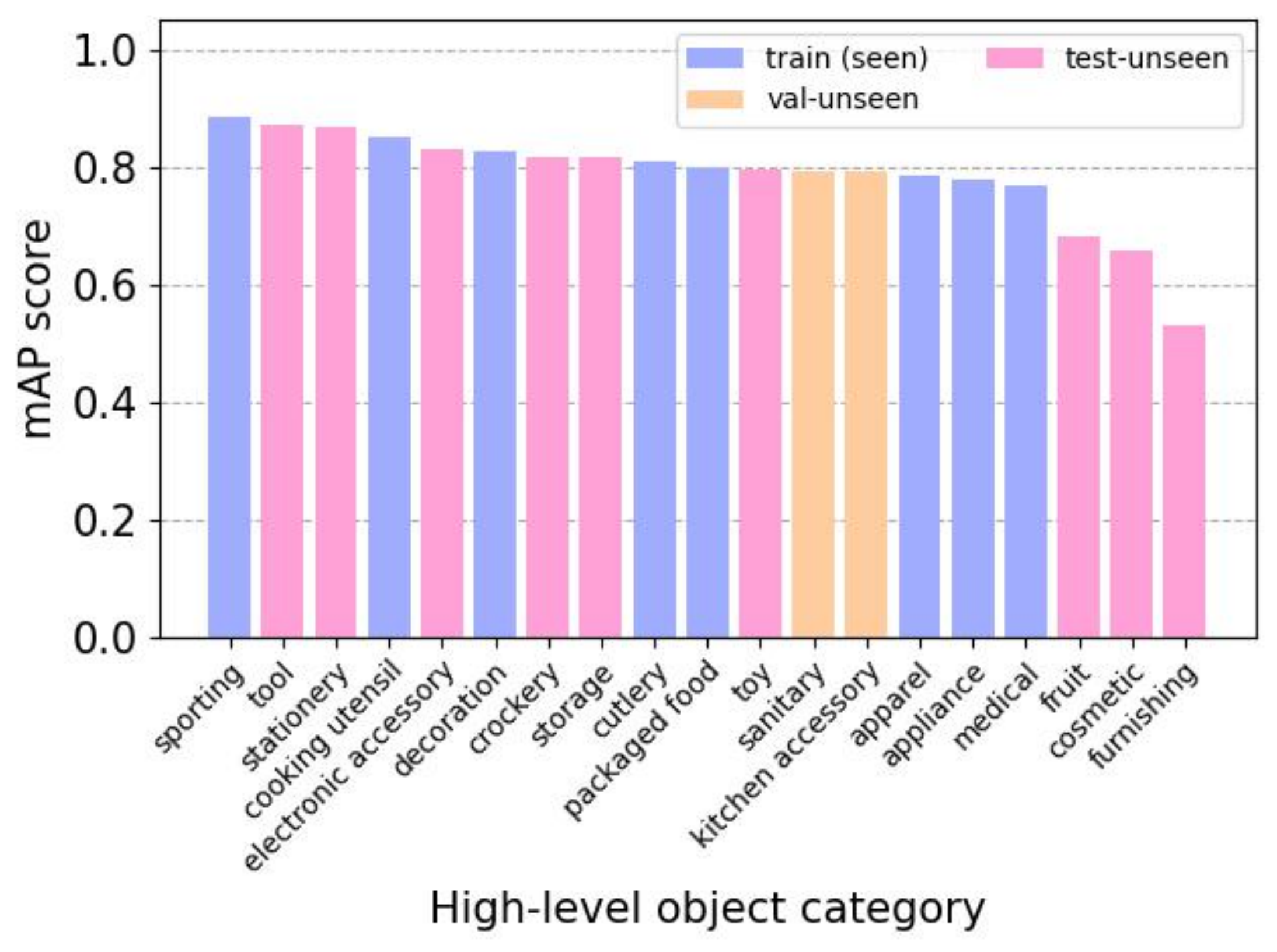}
    \caption{\small{\texttt{ORR} performance of \texttt{RoBERTa + CM} across different high-level categories}}
    \label{fig:orr_per_high_level}
\end{figure}

We now analyze the performance of our \texttt{RoBERTa+CM} scoring function across different high-level categories. We compute mAP scores for \texttt{OR} and \texttt{ORR} tasks (as in Section~\ref{sec:exp:llm}) and average them per high-level object category. While the scoring function performs perfectly (mAP=1) on seen categories for the \texttt{OR} task, the \texttt{OR} task performance drops for unseen high-level categories categories (Figure~\ref{fig:or_per_high_level}). In contrast, the mAP score is close to 0.8 for most seen and unseen high-level categories (Figure~\ref{fig:orr_per_high_level}). The test-unseen high-level categories of fruit, furnishing and cosmetic have low mAP scores for both \texttt{OR} and \texttt{ORR} tasks.

\subsection{Generalization to unseen categories}
In Table~\ref{tab:main:unseen}, we observed that the Object Success on unseen categories when using the language model-based ranking function is comparable Object Success on seen categories. We now provide qualitative examples showing the performance of our \texttt{OR} and \texttt{ORR} scoring functions on unseen categories. 

Figure~\ref{fig:or_qual} shows the ranked list of rooms obtained for each object category using our \texttt{OR} ranking function. We also indicate if the room is a valid room for the given object. Recall that a room is considered valid if it contains at least one receptacle that is deemed \texttt{correct} by at least 6/10 annotators. While the ranked lists for scissors (a tool) and large marker (stationery) have the valid rooms on top, a few valid rooms are further down in the list for banana (fruit category).

Figure~\ref{fig:orr_qual} shows the ranked list of receptacles with the room for the given object-room pair. These ranked lists are obtained using the \texttt{ORR} ranking function. We indicate if the receptacle is a valid receptacle next to the receptacle's name. For the shown examples, most of the valid receptacles are on top of the ranked lists.

\begin{figure}
\scriptsize
    \centering
    \begin{subfigure}[t]{0.3\columnwidth}
    \captionsetup{width=0.95\textwidth}
    \caption{\small{\textbf{Category: }scissors}}
    \begin{tabular}[width=0.95\textwidth]{clc}
        \toprule
        \# & \textbf{Ranked list} & \textbf{Valid?} \\
        \midrule
        1 & kitchen & \cmark \\
        2 & closet & \cmark \\
        3 & playroom & \xmark \\
        4 & utility room & \cmark \\
        5 & dining room & \cmark \\
        6 & bedroom & \cmark \\
        7 & home office & \cmark \\
        8 & garage & \cmark \\
        9 & childs room & \cmark \\
        10 & pantry room & \cmark \\
        11 & bathroom & \cmark \\
        12 & living room & \cmark \\
        13 & television room & \cmark \\
        14 & lobby & \cmark \\
        15 & corridor & \cmark \\
        16 & storage room & \cmark \\
        17 & exercise room & \xmark \\
        \bottomrule
        \end{tabular}
    \end{subfigure}
    \begin{subfigure}[t]{0.3\columnwidth}
    \captionsetup{width=0.95\textwidth}
    \caption{\small{\textbf{Category: }large marker}}
        \begin{tabular}{clc}
            \toprule
            \# & \textbf{Ranked list} & \textbf{Valid?} \\
            \midrule
            1 & closet & \cmark \\
            2 & kitchen & \cmark \\
            3 & garage & \cmark \\
            4 & utility room & \cmark \\
            5 & corridor & \cmark \\
            6 & bedroom & \cmark \\
            7 & dining room & \cmark \\
            8 & childs room & \cmark \\
            9 & playroom & \cmark \\
            10 & television room & \cmark \\
            11 & storage room & \cmark \\
            12 & home office & \cmark \\
            13 & living room & \cmark \\
            14 & pantry room & \xmark \\
            15 & bathroom & \cmark \\
            16 & lobby & \cmark \\
            17 & exercise room & \xmark \\
            \bottomrule
        \end{tabular}
    \end{subfigure}
    \begin{subfigure}[t]{0.3\columnwidth}
    \captionsetup{width=0.95\textwidth}
    \caption{\small{\textbf{Category: }banana}}
    \begin{tabular}{clc}
        \toprule
        \# & \textbf{Ranked list} & \textbf{Valid?} \\
        \midrule
        1 & kitchen & \cmark \\
        2 & garage & \xmark \\
        3 & utility room & \xmark \\
        4 & closet & \xmark \\
        5 & dining room & \xmark \\
        6 & bedroom & \xmark \\
        7 & childs room & \cmark \\
        8 & pantry room & \xmark \\
        9 & home office & \cmark \\
        10 & storage room & \xmark \\
        11 & living room & \xmark \\
        12 & bathroom & \xmark \\
        13 & television room & \xmark \\
        14 & corridor & \xmark \\
        15 & playroom & \xmark \\
        16 & lobby & \xmark \\
        17 & exercise room & \xmark \\
        \bottomrule
        \end{tabular}
    \end{subfigure}
    \caption{\small{OR performance for unseen categories}}
    \label{fig:or_qual}
\end{figure}

\begin{figure}
\scriptsize
    \centering
    \begin{subfigure}[t]{0.3\columnwidth}
    \captionsetup{width=0.95\textwidth}
    \caption{\small{\textbf{Category: }scissors\\\textbf{Room: }living room}}
    \begin{tabular}[width=0.95\textwidth]{clc}
        \toprule
        \# & \textbf{Ranked list} & \textbf{Valid?} \\
        \midrule
        1 & bottom cabinet & \cmark \\
        2 & shelf & \xmark \\
        3 & chest & \cmark \\
        4 & console table & \xmark \\
        5 & table & \xmark \\
        6 & coffee table & \xmark \\
        7 & stool & \xmark \\
        8 & loudspeaker & \xmark \\
        9 & office chair & \xmark \\
        10 & sofa & \xmark \\
        11 & chair & \xmark \\
        12 & speaker system & \xmark \\
        13 & sofa chair & \xmark \\
        14 & carpet & \xmark \\
        \bottomrule
        \end{tabular}
    \end{subfigure}
    \begin{subfigure}[t]{0.3\columnwidth}
    \captionsetup{width=0.95\textwidth}
    \caption{\small{\textbf{Category: }large marker\\\textbf{Room: }corridor}}
        \begin{tabular}{clc}
            \toprule
            \# & \textbf{Ranked list} & \textbf{Valid?} \\
            \midrule
            1 & shelf & \cmark \\
            2 & chest & \cmark \\
            3 & washer & \xmark \\
            4 & console table & \xmark \\
            5 & table & \xmark \\
            6 & dryer & \xmark \\
            7 & chair & \xmark \\
            8 & carpet & \xmark \\
            \bottomrule
        \end{tabular}
    \end{subfigure}
    \begin{subfigure}[t]{0.3\columnwidth}
    \captionsetup{width=0.95\textwidth}
    \caption{\small{\textbf{Category: }banana\\\textbf{Room: }kitchen}}
    \begin{tabular}{clc}
        \toprule
        \# & \textbf{Ranked list} & \textbf{Valid?} \\
        \midrule
        1 & shelf & \cmark \\
        2 & top cabinet & \xmark \\
        3 & bottom cabinet & \xmark \\
        4 & chest & \xmark \\
        5 & counter & \cmark \\
        6 & fridge & \xmark \\
        7 & oven & \xmark \\
        8 & coffee machine & \xmark \\
        9 & sink & \xmark \\
        10 & stove & \xmark \\
        11 & table & \xmark \\
        12 & cooktop & \xmark \\
        13 & carpet & \xmark \\
        14 & dishwasher & \xmark \\
        15 & chair & \xmark \\
        16 & microwave & \xmark \\
        \bottomrule
        \end{tabular}
    \end{subfigure}
    \caption{\small{ORR performance for unseen categories}}
    \label{fig:orr_qual}
\end{figure}

\clearpage

\end{document}